\documentclass[lettersize,journal]{IEEEtran}
\usepackage{amsmath,amsfonts}
\usepackage{algorithmic}
\usepackage{algorithm}
\usepackage{subcaption}
\usepackage{array}
\usepackage[caption=false,font=normalsize,labelfont=sf,textfont=sf]{subfig}
\usepackage{textcomp}
\usepackage{stfloats}
\usepackage{url}
\usepackage{verbatim}
\usepackage{graphicx}
\usepackage{cite}
\begin{document}

\title{Modeling Feature Maps for Quantum Machine Learning}

\author{Navneet Singh 
        and
        Shiva Raj Pokhrel, \textit{IEEE, Senior Member}
        \thanks{N.~Singh and S.~R.~Pokhrel are with Deakin University, Geelong, VIC, Australia. Email: n.navneetsingh@deakin.edu.au; shiva.pokhrel@deakin.edu.au.}}%


\markboth{Journal of \LaTeX\ Class Files,~Vol.~14, No.~8, August~2025}%
{Shell \MakeLowercase{\textit{et al.}}: A Sample Article Using IEEEtran.cls for IEEE Journals}


\maketitle

\begin{abstract}
Quantum Machine Learning (QML) offers significant potential for complex tasks like genome sequence classification, but quantum noise on Noisy Intermediate-Scale Quantum (NISQ) devices poses practical challenges. This study systematically evaluates how various quantum noise models—including dephasing, amplitude damping, depolarizing, thermal noise, bit-flip, and phase-flip—affect key QML algorithms (QSVC, Peg-QSVC, QNN, VQC) and feature mapping techniques (ZFeatureMap, ZZFeatureMap, and PauliFeatureMap). Results indicate that QSVC is notably robust under noise, whereas Peg-QSVC and QNN are more sensitive, particularly to depolarizing and amplitude-damping noise. The PauliFeatureMap is especially vulnerable, highlighting difficulties in maintaining accurate classification under noisy conditions. These findings underscore the critical importance of feature map selection and noise mitigation strategies in optimizing QML for genomic classification, with promising implications for personalized medicine.
\end{abstract}


\section{Introduction}
\IEEEPARstart{T}{he} global quantum technology market is forecasted to reach approximately USD 173 billion by 2040, with quantum computing projected to contribute between USD 28 billion and USD 72 billion by 2035, expanding further to USD 45 billion to USD 131 billion by 2040. This projected growth underscores the importance of applications like Quantum Machine Learning (QML), which are expected to be significant drivers of early quantum computing value creation \cite{alexeev2024quantum}.

In this evolving landscape, QML stands out as a transformative intersection of quantum algorithms and classical machine learning, offering solutions to highly complex problems. Enabled by advances in quantum hardware and algorithms, QML has the potential to revolutionize fields such as data analysis, optimization, and predictive modeling \cite{peral2024systematic,kosoglu2023biological}. QML represents a near-term opportunity for quantum computers to showcase computational advantages over classical approaches, particularly through parameterized quantum circuits \cite{benedetti2019parameterized}, enabling efficient processing of high-dimensional data that remains challenging for traditional methods \cite{domingo2023taking}.

A promising application for QML is genomics, where the ability to analyze extensive genetic datasets is vital for understanding gene-disease relationships and advancing personalized treatments. Genomic data, characterized by its high dimensionality and complex structure, poses significant computational demands, often exceeding the capacity of classical algorithms. QML, leveraging the unique quantum properties of superposition and entanglement, can substantially improve data processing and pattern recognition, providing new insights into genetic diseases and enabling tailored therapies \cite{wong2022fast}.

However, the precision required for genomic analysis makes it particularly sensitive to quantum noise, which presents a major barrier to QML applications in this field \cite{domingo2023taking}. Even minimal quantum noise in genomic classification tasks can lead to data misclassification, reducing the reliability of QML models for critical medical applications such as disease diagnosis and treatment development.
Noisy Intermediate-Scale Quantum (NISQ) devices \cite{preskill2018quantum,huang2022benchmarking}, which offer near-term quantum capabilities \cite{zhu2023variation}, hold promise for revolutionizing genome sequencing and analysis through QML \cite{singh2024independent}. Despite their potential, these devices are highly susceptible to environmental interactions, leading to quantum noise that can disrupt and degrade QML performance. Addressing these limitations is crucial to unlocking the full potential of QML for genomic data processing.

To fully understand and unlock the potential of QML algorithms, it is essential to comprehensively understand how they behave under various quantum noise types arising from current hardware limitations and environmental interactions. To address these hardware constraints, methodologies should be designed to ensure compatibility with existing NISQ devices, balancing the requirements of real-world applications with the limitations imposed by NISQ systems.

Current studies often address hardware limitations by employing small or ad hoc datasets, which do not accurately reflect real-world scenarios and thus may obscure the true potential of QML \cite{li2024quantum, borras2023impact,escudero2023assessing,kashif2024investigating}. Furthermore, existing research on quantum noise in QML models typically examines only a limited set of noise types \cite{li2024quantum, borras2023impact,escudero2023assessing}, leaving room for a more in-depth exploration that could enhance the reliability and applicability of QML in practical settings.

In this work, we investigate the impact of quantum noise on various QML algorithms applied to a genomic sequence dataset \cite{grevsova2023genomic}. By incorporating realistic noise models into our simulations, we assess how different types and levels of noise affect learning processes, convergence, and generalization within QML algorithms. Our research aims to provide insights into optimizing QML for genomics, with a focus on applications such as disease detection and protein folding \cite{pyrkov2024complexity}. 
\begin{figure*}[h!]
    \centering
    \includegraphics[scale=0.7]{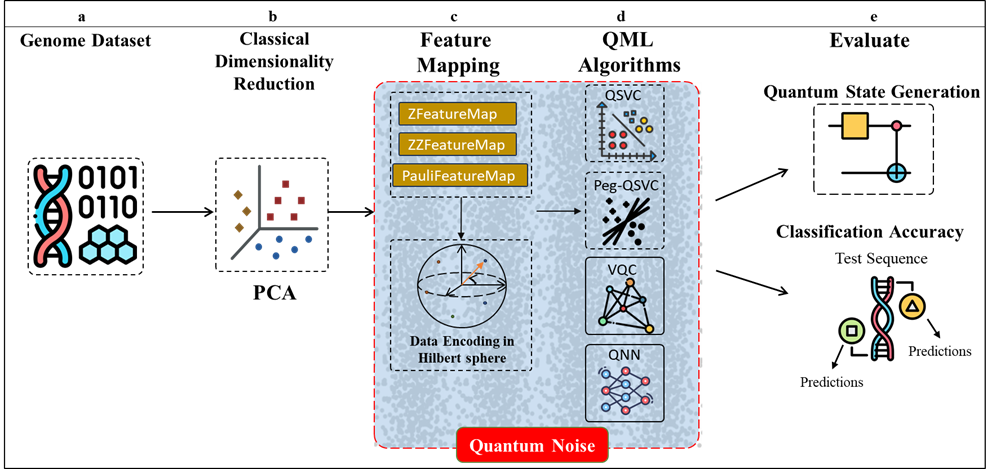}
    \caption{\rm Overview of the proposed workflow presented in this paper. \textbf{a)} Dataset Split: Split the classical dataset into training and testing subsets. \textbf{b)} Classical Dimensionality Reduction: Apply PCA to reduce the dataset to four dimensions. \textbf{c)} Feature Mapping: Transform the dataset into quantum states into in Hilbert space. The performance of these encoding techniques is notably influenced by different types of inherent \textit{quantum noise} in NISQ devices. \textbf{d)} QML Algorithm: Trains various QML algorithms on the encoded quantum data, with \textit{quantum noise} affecting the training process and algorithm performance. \textbf{e)} Evaluation: Assess the impact of \textit{quantum noise} on encoding by generating quantum states and using the trained QML models to classify test sequences.}
    \label{fig:main_flow}
\end{figure*}

We define the binary classification task over a genomic dataset as 
\(\mathcal{D} = \{(\mathbf{x}_i, y_i)\}_{i=1}^N\), where \(\mathbf{x}_i \in \mathbb{R}^d\) represents genome sequences and \(y_i \in \{0, 1\}\) indicates transcript types \cite{grevsova2023genomic} and the basic flow of the work is shown in Fig. \ref{fig:main_flow}. Our QML model \(\mathcal{M}_\theta\) with parameters \(\theta\) aims to predict \(\hat{y}_i = \mathcal{M}_\theta(\mathcal{N}(\mathbf{x}_i))\), where \(\mathcal{N}\) represents quantum noise affecting the genomic data encoded as quantum states. The objective is to minimize the loss function \(\mathcal{L}(\theta)\), defined as:
\(\theta^* = \arg\min_{\theta} \mathbb{E}_{(\mathbf{x}_i, y_i) \sim \mathcal{D}} \left[\mathcal{L}(y_i, \mathcal{M}_\theta(\mathcal{N}(\rho_i)))\right]\).
Model performance is evaluated by classification accuracy, \(\frac{1}{N} \sum_{i=1}^N \mathbb{I}(\hat{y}_i = y_i),\) providing insight into how well the QML model generalizes under quantum noise, where \(\mathbb{I}(\cdot)\) is the indicator function.

\vspace{-4 mm}
\subsection{Our Contributions}
\begin{enumerate}
    \item \textit{Systematic Evaluation of Quantum Noise Effects on redesigning QML Algorithms:} Unlike previous works that focused on specific noise models, we provide a comprehensive analysis of the impact of various quantum noise models such as dephasing, amplitude damping, depolarizing, thermal relaxation noise, bit-flip and phase-flip errors on the performance of QML algorithms applied to genomic data, offering a more holistic understanding of noise effects. By incorporating realistic noise models, our paper highlights how different types and levels of quantum noise influence learning outcomes, convergence, and model generalization.
    
    \item \textit{Optimizing QML for Genomic Data Classification:} We identify critical factors, including feature map selection, that significantly affect the robustness of QML models in genomic sequence classification. These insights advance the development of more noise-resistant QML models, particularly in areas such as disease detection and personalized medicine, where the complexity and precision of genomic data are crucial.

    \item \textit{Practical Insights into the Use of NISQ Devices for Genomic Analysis:} We bridge the gap between theoretical quantum computing and its practical application in genomics by demonstrating how NISQ devices can be used for large-scale genomic data processing. Our study underscores the potential of NISQ-era quantum computers in accelerating tasks such as genome sequencing while also addressing the key challenges posed by quantum noise.
\end{enumerate}
\vspace{-3 mm}
\section{Understanding Quantum Noise}
Developing QML methods capable of operating within realistic, noisy quantum environments relies on both understanding and accurately modeling quantum noise. This section provides an overview of quantum noise and key noise models, illustrating the importance of a comprehensive approach to studying quantum noise.

Quantum noise is an intrinsic aspect of quantum systems, resulting from fundamental interactions with the environment and the probabilistic nature of quantum mechanics. Unlike classical systems, where noise may affect only part of the system, quantum systems are highly sensitive to environmental interactions. These interactions disturb qubit states, causing errors and a loss of coherence. This degradation of quantum coherence leads to non-deterministic errors that are particularly challenging to mitigate \cite{clerk2010introduction}. 
Maintaining qubit coherence over time is essential for efficient quantum computing. Quantum noise introduces errors that can significantly reduce the performance of quantum algorithms. This is especially problematic in QML, where subtle quantum phenomena are essential for data processing and learning. Quantum noise may distort learning outcomes, reducing the effectiveness of QML models. A thorough examination of quantum noise is therefore essential for the design of robust quantum technologies, as noise-induced disturbances can severely impair the performance of QML.

The effects of quantum noise are typically analyzed using a mathematical framework that models the evolution of quantum states under noise. One widely used approach is the Kraus operator formalism \cite{silva2014pre,escher2011quantum}, which characterizes quantum noise as a quantum channel \cite{schlimgen2021quantum}. This quantum channel is a completely positive, trace-preserving map that transforms an initial quantum state into a final state through environmental interaction. The evolution of a quantum state \(\rho\) under noise is given by,
        \(\rho' = \sum_i K_i \rho K_i^\dagger\)
where \(\rho\) is the initial density matrix, \(K_i\) are the Kraus operators satisfying the completeness relation \(\sum_i K_i^\dagger K_i = I\) (with \(I\) as the identity operator), and \(\rho'\) represents the final state after the noise process. Modeling quantum noise with Kraus operators enables accurate representation of various noise types, offering a systematic approach for assessing how different noise models impact QML efficiency.

The following subsections provide a detailed overview of several types of quantum noise, including their mathematical modeling, an intuitive understanding, and the implications of each for QML.
\vspace{-3 mm}
\subsection{Dephasing/ Phase Damping Noise}
Dephasing noise, also referred to as phase damping\footnote{The terms dephasing and phase damping noise are used interchangeably.}, weakens a quantum system's phase coherence while leaving populations and energy levels unchanged. This type of noise arises when environmental interactions cause the qubit’s phase to fluctuate randomly, creating uncertainty about its coherence relative to other quantum states \cite{crow2014classical}. Over time, this interaction leads to a gradual loss of quantum coherence \cite{xue2021effects}. In practical quantum systems, dephasing typically results from fluctuating electromagnetic fields, thermal noise, or similar external perturbations affecting qubit energy levels.

Using the Kraus operator formalism, dephasing noise affecting a single qubit is modeled with the following operators, 
$K_0 = \sqrt{1 - p}I$,  $K_1 = \sqrt{p} \, \sigma_z$, 
where \(p\) is the probability of a phase flip occurring, \(I\) is the identity matrix, and \(\sigma_z\) is the Pauli-Z operator.
The evolution of the qubit's density matrix \(\rho\) under dephasing noise is given by, $\rho' = K_0 \rho K_0^\dagger + K_1 \rho K_1^\dagger$ and by substituting the Kraus operators into it, we obtain $\rho' = (1 - p) \rho + p \, \sigma_z \rho \sigma_z$. For an initial density matrix \(\rho\) expressed and the evolution under dephasing noise simplifies to, $\rho = \begin{pmatrix} \rho_{00} & \rho_{01} \\ \rho_{10} & \rho_{11} \end{pmatrix} = \begin{pmatrix} \rho_{00} & (1 - 2p) \rho_{01} \\ (1 - 2p) \rho_{10} & \rho_{11} \end{pmatrix}$. 
The off-diagonal elements (coherence terms) of \(\rho\) are reduced by a factor of \((1 - 2p)\), while the diagonal elements (populations) remain unchanged \cite{chen2021low}. In QML, qubit coherency is essential for quantum parallelism and interference effects, which give quantum computers their full computational capabilities. Dephasing noise typically lowers the quantum coherence required for these quantum events, which has a significant effect on how well those quantum models operate. For instance, dephasing causes an information loss in quantum classifier models due to the loss of relative phases encoded in the qubits, which are essential for accurate prediction and learning.
We first investigate the effects of dephasing noise on qubit phase coherence, and then we focus on amplitude damping noise, which results in a qubit losing its energy state to the environment.

\vspace{-4 mm}

\subsection{Amplitude damping noise}
Amplitude damping noise models the loss of energy from a quantum system to its environment, particularly relevant in systems experiencing spontaneous emission or energy relaxation. It describes how a qubit decays from an excited state \(|1\rangle\) to a ground state \(|0\rangle\) by emitting a photon or other excitations \cite{srikanth2008squeezed}. It is crucial as it represents a fundamental mechanism by which quantum information is lost due to energy dissipation.
Amplitude damping occurs naturally due to interactions with the electromagnetic field or thermal environments, where energy exchange leads to the relaxation of excited states.
The evolution of a quantum system under amplitude damping noise is described using the Kraus operators, 
$K_0 = \begin{pmatrix} 1 & 0 \\ 0 & \sqrt{1 - \gamma} \end{pmatrix}$, $K_1 = \begin{pmatrix} 0 & \sqrt{\gamma} \\ 0 & 0 \end{pmatrix}$.
Here, \(\gamma\) is the probability of the qubit decaying from the excited state \(|1\rangle\) to the ground state \(|0\rangle\) \cite{rodriguez2023suppressing}.
The evolved density matrix is,
$\rho' = \begin{pmatrix} \rho_{00} + \gamma \rho_{11} & \sqrt{1 - \gamma} \, \rho_{01} \\ \sqrt{1 - \gamma} \, \rho_{10} & (1 - \gamma) \rho_{11} \end{pmatrix}$.
This indicates that the population in the excited state \(\rho_{11}\) decreases by a factor of \((1 - \gamma)\), reflecting the system's relaxation to the ground state. The coherence terms \(\rho_{01}\) and \(\rho_{10}\) are reduced by \((\sqrt{1 - \gamma})\).
Amplitude damping noise affects QML algorithms by altering the populations of qubit states and reducing coherence. In quantum circuits used for machine learning tasks, such as variational algorithms or quantum feature maps, amplitude damping can lead to a loss of information encoded in the amplitude of the quantum states. This may result in decreased accuracy of the learning model or convergence issues during training. 
While amplitude damping models energy loss, another important type of noise is depolarizing noise, which represents random errors affecting both the amplitude and phase of qubits. We explore this next.
\vspace{-3mm}
\subsection{Depolarizing Noise}
Depolarizing noise is a type of quantum noise that describes the process where a quantum system loses its coherence by becoming a maximally mixed state \cite{imany2019high}. It models the effect of random errors that can flip the qubit's state or change its phase, effectively depolarizing the quantum state. Depolarizing noise can result from various imperfections in quantum hardware, such as gate errors, cross-talk between qubits, or uncontrolled interactions with the environment.
The Kraus operators for depolarizing noise \cite{gokhale2019asymptotic} are, 
$K_0 = \sqrt{1 - \frac{3p}{4}}I$, $K_1 = \sqrt{\frac{p}{4}}X$, $K_2 = \sqrt{\frac{p}{4}}Y$, $K_3 = \sqrt{\frac{p}{4}}Z$. 
Here, \(p\) is the depolarizing probability, and \(X\), \(Y\), and \(Z\) are the Pauli matrices and the identity matrix \(I\). 
The density matrix \(\rho\) of a qubit evolves under depolarizing noise according to the Kraus operators as, \(\rho' = \sum_{i=0}^{3} K_i \rho K_i^\dagger\).
Under depolarizing noise, the density matrix, $\rho$ evolves as, \( \rho' = \begin{pmatrix} (1 - p) \rho_{00} + \dfrac{p}{2} & (1 - p) \rho_{01} \\ (1 - p) \rho_{10} & (1 - p) \rho_{11} + \dfrac{p}{2} \end{pmatrix} \).
This expression shows that the effect of depolarizing noise is to shrink the off-diagonal coherence terms \(\rho_{01}\) and \(\rho_{10}\) by a factor of \((1 - p)\), leading to a reduction in quantum coherence. Additionally, the populations \(\rho_{00}\) and \(\rho_{11}\) are driven towards \(\frac{1}{2}\), corresponding to the maximally mixed state. This result implies that the quantum state loses its distinguishability, effectively becoming more random as the noise increases.

Depolarizing noise is particularly detrimental to QML algorithms because it represents a complete loss of quantum information. As QML models often rely on subtle quantum correlations and superpositions to encode and process information, depolarizing noise can severely impair their functionality. In addition to energy loss and random errors, quantum systems are also affected by thermal relaxation due to their interaction with thermal environments. We examine thermal relaxation noise next.

\vspace{-4 mm}
\subsection{Thermal Relaxation Noise}

Thermal relaxation noise models the effects of a qubit interacting with a thermal environment, causing transitions between energy states due to absorption and emission of thermal energy. This includes both relaxation from excited states to ground states and excitation from ground states to excited states due to thermal energy.
In a finite-temperature environment, qubits can absorb thermal energy, leading to excitation, or release energy, leading to relaxation. This bidirectional energy exchange is characterized by thermal relaxation time and depends on the temperature of the environment \cite{rost2020simulation}.

The Kraus operators for thermal relaxation noise are,
$K_0 = \sqrt{1 - p_0 - p_1}I$, $K_1 = \sqrt{p_1}\sigma_-$, $K_2 = \sqrt{p_0}\sigma_+$. Here, \(p_0\) and \(p_1\) are the probabilities associated with thermal excitation and relaxation processes, respectively, and are related to the relaxation time \(T_1\) and the temperature of the environment \cite{tolunay2023hamiltonian}. The operators $\sigma_-$ and $\sigma_+$ are the lowering and raising operators.
The density matrix \(\rho\) of a qubit under the influence of thermal noise evolves according to the Kraus operators as,
$\rho' = \sum_{i=0}^{2} K_i \rho K_i^\dagger$, and the resulting evolved density matrix is, $\rho' = (1 - p_0 - p_1) \rho + p_1 \sigma_- \rho \sigma_-^\dagger + p_0 \sigma_+ \rho \sigma_+^\dagger$. After expanding the terms, we have get,
$\rho' = \begin{pmatrix} (1 - p_0 - p_1) \rho_{00} + p_0 \rho_{11} & (1 - p_0 - p_1) \rho_{01} \\ (1 - p_0 - p_1) \rho_{10} & (1 - p_0 - p_1) \rho_{11} + p_1 \rho_{00} \end{pmatrix}$. 

Thermal relaxation noise redistributes the qubit populations and reduces coherence. The probability of the ground state \(\rho_{00}\) increases by \(p_0 \rho_{11}\), while the excited state \(\rho_{11}\) decreases by \((1 - p_0 - p_1)\) but gains \(p_1 \rho_{00}\). The off-diagonal coherence terms \(\rho_{01}\) and \(\rho_{10}\) are diminished by \((1 - p_0 - p_1)\), reflecting a loss of coherence and energy exchanges with the environment.

Thermal relaxation noise affects QML algorithms by altering the energy populations of qubits in a way that depends on the environment's temperature. This can introduce errors in computations and reduce the fidelity of quantum operations. For instance, in algorithms that require precise control of qubit states, thermal excitations can introduce unwanted transitions that lead to incorrect results. 
In addition to continuous noise processes, discrete errors such as bit-flip and phase-flip errors also play a significant role in quantum computations. We now discuss these errors and their impact on quantum algorithms.

\vspace{-4 mm}
\subsection{Bit-Flip Errors}

Bit-flip errors occur when a qubit's state is flipped from \(|0\rangle\) to \(|1\rangle\) or vice versa. This type of error is analogous to classical bit-flip errors in digital systems but has quantum mechanical origins, such as interactions with the environment that cause energy transitions or imperfections in quantum gates  \cite{funcke2022measurement}. 
For a single qubit, the Kraus operators that model bit-flip noise are \cite{jiang2022teleportation}, 
$E_0 = \sqrt{1 - p}I$, $E_1 = \sqrt{p}X$, where \(I\) is the identity operator, \(X\) is the Pauli-X operator, and \(p\) is the probability of a bit-flip occurring.
The action of the bit-flip channel on the density matrix \(\rho\) is,
$\rho' = E_0 \rho E_0^\dagger + E_1 \rho E_1^\dagger$, and expanding it gives us, $\rho' = (1 - p) \rho + p X \rho X$.
This equation indicates that with probability \((1 - p)\), the state remains unchanged, and with probability \(p\), the state undergoes a bit-flip.
For a pure state $|\psi\rangle = \alpha |0\rangle + \beta |1\rangle$, the corresponding density matrix is,
$\rho = |\psi\rangle \langle \psi| = \begin{pmatrix}|\alpha|^2 & \alpha \beta^* \\ \alpha^* \beta & |\beta|^2\end{pmatrix}$
After applying the bit-flip channel and by expanding, the evolved density matrix becomes, $\rho' = \begin{pmatrix}
(1 - p)|\alpha|^2 + p|\beta|^2 & (1 - p)\alpha \beta^* + p\alpha^* \beta \\
(1 - p)\alpha^* \beta + p\alpha \beta^* & (1 - p)|\beta|^2 + p|\alpha|^2
\end{pmatrix}$
This shows that the bit-flip noise redistributes the population between \(|0\rangle\) and \(|1\rangle\), affecting the coherence terms by reducing the off-diagonal elements. 

Bit-flip errors can introduce significant inaccuracies in QML algorithms by changing the logical state of qubits. For example, in quantum circuits implementing machine learning models, bit-flip errors can lead to incorrect encoding of data or erroneous computation results.

\vspace{-4 mm}
\subsection{Phase-Flip Errors}

Phase-flip errors affect the relative phase between quantum states without altering their populations. Specifically, they change the sign of the phase of the \(|1\rangle\) state, effectively flipping the phase of the qubit \cite{jiang2022teleportation}. This type of error disrupts the interference patterns essential for many quantum algorithms. This noise affects the relative phase between the basis states but leaves their amplitudes unchanged.  Phase-flip errors are often considered alongside bit-flip errors, as they can occur due to similar environmental interactions, but their impact is on the phase rather than the amplitude of the quantum state.

The Kraus operators for phase-flip noise \cite{jiang2022teleportation} are,$E_0 = \sqrt{1 - p}I, \quad E_1 = \sqrt{p}Z$
where \(I\) is the identity operator and \(Z\) is the Pauli-Z operator.
Here, \(p\) is the probability of a phase-flip occurring. The action on the density matrix \(\rho\) is described by,
$\rho' = E_0 \rho E_0^\dagger + E_1 \rho E_1^\dagger = (1 - p) \rho + p Z \rho Z$
This equation indicates that with probability \(1 - p\), the state remains unchanged, and with probability \(p\), the phase of the state undergoes a flip.
After applying the phase-flip channel and simplifying the expression, the resulting evolved density matrix for a pure state is given by, $\rho' = \begin{pmatrix}
|\alpha|^2 & (1 - 2p)\alpha \beta^* \\
(1 - 2p)\alpha^* \beta & |\beta|^2
\end{pmatrix}$. 
Phase-flip errors can degrade the performance of QML algorithms by disrupting quantum superpositions and entanglement, which rely on precise phase relationships. In QML, phase-flip errors can lead to incorrect interference outcomes, affecting the learning process.

\begin{table*}[ht!]
\caption{Comparison of Different Quantum Noise Types}
\label{tab:my-table}
\resizebox{\textwidth}{!}{%
\begin{tabular}{|c|c|c|c|c|c|c|}
\hline
\textbf{Aspect} & \textbf{\begin{tabular}[c]{@{}c@{}}Bit-Flip\\ Noise\end{tabular}} & \textbf{\begin{tabular}[c]{@{}c@{}}Phase-Flip\\ Noise\end{tabular}} & \textbf{\begin{tabular}[c]{@{}c@{}}Depolarizing \\ Noise\end{tabular}} & \textbf{\begin{tabular}[c]{@{}c@{}}Dephasing \\ Noise\end{tabular}} & \textbf{\begin{tabular}[c]{@{}c@{}}Amplitude \\ Damping Noise\end{tabular}} & \textbf{\begin{tabular}[c]{@{}c@{}}Thermal Relaxation\\ Noise\end{tabular}} \\ \hline
\textbf{Nature} & \begin{tabular}[c]{@{}c@{}}Flips qubit state: \\ \(\lvert 0 \rangle \leftrightarrow \lvert 1 \rangle\)\end{tabular} & \begin{tabular}[c]{@{}c@{}}Flips phase: \\ \(\lvert 1 \rangle \to -\lvert 1 \rangle\)\end{tabular} & \begin{tabular}[c]{@{}c@{}}Randomly flips state \\ in all directions\end{tabular} & \begin{tabular}[c]{@{}c@{}}Loss of coherence\\ in superpositions\end{tabular} & \begin{tabular}[c]{@{}c@{}}Energy loss: \\ \(\lvert 1 \rangle \to \lvert 0 \rangle\)\end{tabular} & \begin{tabular}[c]{@{}c@{}}Energy exchange \\ with the environment\end{tabular} \\ \hline
\textbf{\begin{tabular}[c]{@{}c@{}}Effect on Density \\ Matrix\end{tabular}} & \begin{tabular}[c]{@{}c@{}}Changes diagonal \\ elements based on \\ flip probability\end{tabular} & \begin{tabular}[c]{@{}c@{}}Changes diagonal \\ elements based on \\ flip probability\end{tabular} & \begin{tabular}[c]{@{}c@{}}Alters both diagonal \\ and off-diagonal \\ elements\end{tabular} & \begin{tabular}[c]{@{}c@{}}Decays off-diagonal \\ elements\end{tabular} & \begin{tabular}[c]{@{}c@{}}Reduces amplitude of \\ \(\lvert 1 \rangle\) component\end{tabular} & \begin{tabular}[c]{@{}c@{}}Redistributes \\ population \\ between states, \\ reduces coherence\end{tabular} \\ \hline
\textbf{\begin{tabular}[c]{@{}c@{}}Mathematical \\ Model\end{tabular}} & \(\rho' = (1 - p) \rho + p X \rho X\) & \(\rho' = (1 - p) \rho + p Z \rho Z\) & \begin{tabular}[c]{@{}c@{}}\(\rho' = (1 - p) \rho\) \\ + \(\frac{p}{3} (X \rho X\) \\ + \(Y \rho Y + Z \rho Z)\)\end{tabular} & \(\rho' = (1 - p) \rho + p Z \rho Z\) & \begin{tabular}[c]{@{}c@{}}\(\rho' = E_0 \rho E_0^\dagger + E_1 \rho E_1^\dagger\), \\ \(E_0 = \begin{pmatrix} 1 & 0 \\ 0 & \sqrt{1 - p} \end{pmatrix}\), \\ \(E_1 = \begin{pmatrix} 0 & \sqrt{p} \\ 0 & 0 \end{pmatrix}\)\end{tabular} & \begin{tabular}[c]{@{}c@{}}\(\rho' = (1 - p_0 - p_1) \rho\) \\ + \(p_1 \sigma_- \rho \sigma_-^\dagger\) \\ + \(p_0 \sigma_+ \rho \sigma_+^\dagger\)\end{tabular} \\ \hline
\textbf{\begin{tabular}[c]{@{}c@{}}Impact on \\ Quantum State\end{tabular}} & \begin{tabular}[c]{@{}c@{}}Probability \(p\) of \\ state flipping\end{tabular} & \begin{tabular}[c]{@{}c@{}}Probability \(p\) of \\ phase flipping\end{tabular} & \begin{tabular}[c]{@{}c@{}}Probability \(p\) of \\ complete depolarization\end{tabular} & \begin{tabular}[c]{@{}c@{}}Coherence loss over \\ time, state remains \\ pure if no \\ interaction\end{tabular} & \begin{tabular}[c]{@{}c@{}}Probability \(p\) of energy \\ relaxation to \(\lvert 0 \rangle\)\end{tabular} & \begin{tabular}[c]{@{}c@{}}Probability of state \\ transition due to \\ thermal excitation\end{tabular} \\ \hline
\textbf{\begin{tabular}[c]{@{}c@{}}Physical \\ Interpretation\end{tabular}} & \begin{tabular}[c]{@{}c@{}}Models errors in \\ classical bit-flips\end{tabular} & \begin{tabular}[c]{@{}c@{}}Phase errors, \\ common in \\ superconducting \\ qubits\end{tabular} & \begin{tabular}[c]{@{}c@{}}Models loss of all \\ quantum information\end{tabular} & \begin{tabular}[c]{@{}c@{}}Common in all \\ quantum systems, \\ especially in thermal \\ environments\end{tabular} & \begin{tabular}[c]{@{}c@{}}Represents energy \\ loss due to \\ interaction with \\ the environment\end{tabular} & \begin{tabular}[c]{@{}c@{}}Thermal relaxation \\ due to \\ interaction with \\ environment (e.g., \\ superconducting \\ qubits)\end{tabular} \\ \hline
\textbf{\begin{tabular}[c]{@{}c@{}}Fidelity \\ Impact\end{tabular}} & \begin{tabular}[c]{@{}c@{}}Moderate, \\ increases with \(p\)\end{tabular} & \begin{tabular}[c]{@{}c@{}}Moderate, \\ phase coherence \\ affected\end{tabular} & \begin{tabular}[c]{@{}c@{}}Severe, can result in \\ a complete loss of \\ information\end{tabular} & \begin{tabular}[c]{@{}c@{}}Severe, leads to pure \\ dephasing \\ or full decoherence\end{tabular} & \begin{tabular}[c]{@{}c@{}}High, \\ especially in \\ systems with \\ high \(p\)\end{tabular} & \begin{tabular}[c]{@{}c@{}}High, especially \\ at elevated \\ temperatures\end{tabular} \\ \hline
\end{tabular}%
}
\end{table*}

In summary, a thorough understanding of the various types of quantum noise and their respective effects on quantum systems is critical for the development of reliable and efficient QML algorithms. By employing the Kraus operator formalism to model these noise processes, we can accurately simulate the performance of quantum algorithms in realistic, noisy environments. This knowledge is indispensable for advancing robust quantum technologies and ensuring that QML algorithms remain effective in practical applications. Table \ref{tab:my-table} provides a comprehensive comparison of the quantum noise models discussed in this work, summarizing their characteristics, mathematical representations, and their respective impacts on quantum states. This table highlights both the commonalities and distinctions between the different types of quantum noise. Understanding these effects is crucial for developing robust QML models capable of operating effectively on current and near-future quantum hardware.

\vspace{-4 mm}
\subsection{Feature Mapping in QML} 
In QML, feature mapping plays a pivotal role by encoding classical data into quantum states, allowing quantum algorithms to learn from and process the data efficiently \cite{havlivcek2019supervised}. The choice of feature map can have a profound influence on the performance of a QML model, as it determines how information is represented and manipulated within the quantum system. Different feature maps provide varying levels of expressivity and are more or less resilient to quantum noise, which can affect the quantum states during encoding and processing. This subsection explores three feature mapping methods commonly implemented in Qiskit: the ZFeatureMap, ZZFeatureMap, and PauliFeatureMap.

\subsubsection{ZFeatureMap}

The ZFeatureMap \cite{schuld2020circuit} encodes classical data into quantum states by applying rotations around the Z-axis of the Bloch sphere for each qubit in the quantum register. This feature map is relatively straightforward, making it suitable for datasets where features are independent and do not require interactions between qubits.
In the ZFeatureMap, each classical feature is mapped to a rotation around the Z-axis for the corresponding qubit. This transformation encodes the data in the phase of the quantum state without introducing entanglement between qubits. It is ideal for problems where feature independence is sufficient, but it lacks the ability to capture interactions or correlations between features.
The ZFeatureMap’s lack of entanglement between qubits makes it relatively less sensitive to noise that disrupts qubit interactions. However, phase noise (dephasing) can still affect the rotation angles and alter the quantum state’s phases, potentially leading to inaccuracies in the encoded information.

While the ZFeatureMap is effective for encoding independent features, many real-world datasets exhibit correlations between features that must be captured to achieve accurate predictions. This brings us to the ZZFeatureMap \cite{havlivcek2019supervised}, which enhances the expressivity of the feature map by incorporating entanglement between qubits.

\subsubsection{ZZFeatureMap}

The ZZFeatureMap \cite{havlivcek2019supervised} extends the ZFeatureMap by introducing entanglement between qubits. This feature map is more expressive than the ZFeatureMap because it captures pairwise interactions between features, making it well-suited for problems where feature correlations are important.
While the ZFeatureMap focuses on individual feature rotations, the ZZFeatureMap incorporates entanglement, which allows it to capture relationships between features. By adding entangling gates between qubits, the ZZFeatureMap encodes interactions between pairs of features, providing a richer and more expressive representation of the data. This expressivity comes at the cost of increased sensitivity to quantum noise, particularly errors that affect entangled qubits.
The introduction of entanglement makes the ZZFeatureMap more expressive but also more vulnerable to quantum noise, particularly noise that affects the coherence between qubits. Errors in the entanglement process, such as phase-flip noise, can disrupt the pairwise feature interactions, leading to degraded performance in learning tasks that rely on feature correlations.

The ZZFeatureMap captures pairwise feature interactions, but some datasets require higher-order interactions. The PauliFeatureMap \cite{havlivcek2019supervised,benedetti2019generative} extends this by using multi-axis rotations and complex qubit entanglement to encode more intricate feature dependencies.

\subsubsection{PauliFeatureMap}
The PauliFeatureMap \cite{havlivcek2019supervised,benedetti2019generative} generalizes the feature mapping process by incorporating rotations around multiple axes (X, Y, and Z) and complex entangling operations. This feature map is highly expressive, capable of capturing higher-order interactions and intricate patterns within the data, making it particularly powerful for complex datasets.
The PauliFeatureMap goes beyond simple rotations around the Z-axis by allowing rotations around multiple Pauli axes (X, Y, and Z) and introducing more intricate entangling operations between qubits. This added complexity enables the PauliFeatureMap to encode higher-order interactions between features, making it suitable for datasets where relationships between features are complex and non-linear.

The PauliFeatureMap’s complexity makes it more susceptible to quantum noise. Errors in any of the rotations or entangling operations can lead to significant deviations from the intended quantum state, especially for multi-layer circuits. 

Feature maps play a crucial role in determining the expressivity and noise resilience of QML models. Simpler feature maps like the ZFeatureMap are less expressive but also less vulnerable to noise, while more complex maps like the ZZFeatureMap and PauliFeatureMap provide richer representations at the cost of increased sensitivity to quantum errors. Understanding these trade-offs is essential for choosing the right feature map based on the noise characteristics of the quantum hardware and the complexity of the learning task.

\vspace{-4 mm}
\section{Modeling QML Algorithms}

In this section, we focus on modeling the impact of quantum noise on four representative QML algorithms: i) Quantum Support Vector Classifier (QSVC);
  ii) Pegasos-QSVC (Peg-QSVC);  iii) QNN; iv) Variational Quantum Classifier (VQC). These algorithms are selected due to their foundational roles in QML and their reliance on quantum phenomena like superposition and entanglement, making them particularly susceptible to quantum noise. Understanding how quantum noise affects these algorithms is crucial for advancing QML applications in NISQ devices, particularly in fields like genomics where data complexity demands robust computational methods.

\vspace{-3 mm}
\subsection{QSVC}
The QSVC extends the classical Support Vector Machine (SVM) into the quantum realm, leveraging quantum computing to handle high-dimensional data efficiently. The core idea is to use a quantum feature map to encode classical data into quantum states, enabling the computation of inner products (quantum kernels) in a Hilbert space that may be computationally infeasible classically \cite{rebentrost2014quantum,gentinetta2024complexity}. In QSVC, data is encoded into quantum states and utilizing quantum parallelism to evaluate the kernel function efficiently. The quantum kernel measures the similarity between data points in the quantum feature space.

Quantum noise can disrupt the delicate quantum states used in QSVC, leading to inaccuracies in kernel computations. This affects the classifier's ability to correctly measure the similarity between data points, potentially degrading its classification performance. Modeling the impact of quantum noise allows us to anticipate these issues.
Our goal is to train a QSVC model \(\mathcal{M}_\theta\) with parameters \(\theta\) on the noisy quantum states \(\mathcal{N}(\rho_i)\). QSVC first encodes the classical data \(\mathbf{x}_i \in \mathbb{R}^d\) into quantum states \(\rho_i\). The core of the QSVC relies on the quantum kernel, defined as,$K(\mathbf{x}_i, \mathbf{x}_j) = \text{Tr}[\phi(\mathbf{x}_i) \phi(\mathbf{x}_j)]$,
here \(K(\mathbf{x}_i, \mathbf{x}_j)\) is the quantum kernel function, and \(\text{Tr}[\cdot]\) denotes the trace operation. This kernel measures the similarity between two quantum states corresponding to the input vectors \(\mathbf{x}_i\) and \(\mathbf{x}_j\).

The QSVC model seeks to find a decision function $f(\mathbf{x}) = \text{sign}\left(\sum_{i=1}^N \alpha_i y_i K(\mathbf{x}_i, \mathbf{x}) + b\right)$. Here, $\alpha_i$ is the Lagrange multipliers, optimized during training, \(K(\mathbf{x}_i, \mathbf{x})\) is the quantum kernel between the training data \(\mathbf{x}_i\) and the input \(\mathbf{x}\),  \(b\) is the bias term \cite{gentinetta2024complexity,jager2023universal}. 
When quantum noise is introduced into the system, it affects the quantum states as \(\rho_i' = \mathcal{N}(\rho_i)\). The noisy quantum kernel can be defined as, $K'(\mathbf{x}_i, \mathbf{x}_j) = \text{Tr}[\mathcal{N}(\rho_i) \mathcal{N}(\rho_j)]$.
This  \(K'(\mathbf{x}_i, \mathbf{x}_j)\) is then used in the QSVC decision function, $f'(\mathbf{x}) = \text{sign}\left(\sum_{i=1}^N \alpha_i y_i K'(\mathbf{x}_i, \mathbf{x}) + b\right)$. The goal is to find parameters \(\theta\) that minimize the loss function \(\mathcal{L}(\theta)\) despite noise, $\theta^* = \arg\min_{\theta} \mathbb{E}_{(\mathbf{x}_i, y_i)} \left[\mathcal{L}(y_i, f'(\mathbf{x}_i))\right]$.

While the QSVC is a classical SVM extension that benefits from quantum-enhanced kernels, the Peg-QSVC, discussed next, applies a stochastic gradient descent approach for optimization, improving scalability and efficiency.

\vspace{-3 mm}
\subsection{Peg-QSVC}

The Peg-QSVC combines the stochastic gradient descent (SGD) efficiency of the Pegasos algorithm with quantum computing's capabilities. It aims to solve the SVM optimization problem using quantum kernels, providing a scalable approach to large datasets.
Peg-QSVC iteratively updates the model parameters using SGD, utilizing quantum computations to evaluate the necessary inner products efficiently. The use of quantum kernels allows the algorithm to capture complex patterns in data \cite{shalev2007pegasos}.
Quantum noise impacts the Peg-QSVC by introducing errors in the quantum kernel evaluations during the iterative updates. This can slow down convergence or lead to convergence to suboptimal solutions. Understanding the noise effects is vital for ensuring the reliability and efficiency of the Peg-QSVC.

The goal of Peg-QSVC is to find the optimal parameters $\theta$ that minimize the regularized hinge loss function\cite{havlivcek2019supervised}. The SVM optimization problem for the binary classification task can be expressed as, $\min_{\mathbf{w}, b} \frac{1}{2} \|\mathbf{w}\|^2 + \frac{\lambda}{N} \sum_{i=1}^N \max(0, 1 - y_i(\mathbf{w} \cdot \phi(\mathbf{x}_i) + b))$. Here, $\mathbf{w}$ is the weight vector,  $b$ is the bias term, $\lambda > 0$ is the regularization parameter, $\phi(\mathbf{x}_i)$ is the quantum feature map of the input $\mathbf{x}_i$, which maps the input data to a higher-dimensional quantum Hilbert space, $y_i \in \{0, 1\}$ are the binary labels.
The optimization is performed iteratively by updating $\mathbf{w}$ and $b$ using stochastic sub-gradient descent. For each iteration $t$, a random sample $(\mathbf{x}_i, y_i)$ is selected, and the weight vector is updated as, $\mathbf{w}_{t+1} = \left(1 - \frac{1}{t}\right)\mathbf{w}_t + \frac{1}{\lambda t} y_i \phi(\mathbf{x}_i) \mathbb{I}(y_i(\mathbf{w}_t \cdot \phi(\mathbf{x}_i) + b) < 1)$. 
The objective is to minimize the noisy hinge loss function, which now depends on $\rho'_i$. This is achieved by finding the optimal parameters $\theta^*$ that satisfy,
 $\theta^* = \arg\min_{\theta} \frac{1}{2} \|\mathbf{w}\|^2 + \frac{\lambda}{N} \sum_{i=1}^N \max(0, 1 - y_i(\mathbf{w} \cdot \phi(\mathcal{N}(\rho_i)) + b))$. Here, the optimization process iteratively updates $\mathbf{w}$ by incorporating the noisy quantum feature mapping $\phi(\mathcal{N}(\rho_i))$. The aim is to minimize the loss function while accounting for the effects of quantum noise.
 After examining the noise impact on QSVC-based models, we now turn to the QNN, which employ a fundamentally different approach, leveraging quantum circuits analogous to classical neural networks for learning complex patterns.
 
 \vspace{-4 mm}
\subsection{QNN}

QNNs aim to replicate the success of classical neural networks by using quantum circuits to model complex functions. QNNs leverage quantum superposition and entanglement to achieve higher expressiveness and computational advantages.
A QNN consists of layers of quantum gates with tunable parameters, similar to weights in classical neural networks. The network processes input quantum states and produces outputs through measurements, which are used for predictions \cite{abbas2021power}.
Quantum noise can disrupt delicate quantum states within the QNN, leading to errors in the output probabilities and gradients used for training. Noise can affect the forward propagation of quantum states and the backward propagation of gradients, making training difficult.

A QNN consists of a series of quantum gates arranged in layers, where each layer can be considered analogous to a layer in a classical neural network. The network parameters $\theta$ correspond to the angles of rotation gates or other tunable quantum gates within the circuit \cite{abbas2021power}.
Let $\mathbf{U}(\theta)$ denote the unitary operation representing the entire QNN circuit. The QNN takes a quantum state $\rho_i$ as input.
The output state of the QNN, after applying the unitary operation is given by $\rho_{\text{out}, i} = \mathbf{U}(\theta) \rho_i \mathbf{U}^\dagger(\theta)$, 
where $\mathbf{U}^\dagger(\theta)$ is the Hermitian conjugate of $\mathbf{U}(\theta)$.
The noisy output state is $\rho'_{\text{out}, i} = \mathcal{N}(\rho_{\text{out}, i}) = \mathcal{N}\left(\mathbf{U}(\theta) \rho_i \mathbf{U}^\dagger(\theta)\right)$. The noise can be modeled by various types of quantum channels as disccused above, which introduce different forms of perturbations to the quantum states.
To obtain the predicted label $\hat{y}_i$, a measurement is performed on the noisy output state $\rho'_{\text{out}, i}$. The probability of measuring the state $|1\rangle$ is given by, $P(\hat{y}_i = 1) = \text{Tr}\left(\rho'_{\text{out}, i} |1\rangle \langle 1|\right)$.
Using binary cross-entropy loss, $\mathcal{L}(\theta) = -\frac{1}{N} \sum_{i=1}^N \left[ y_i \log P(\hat{y}_i = 1) + (1 - y_i) \log (1 - P(\hat{y}_i = 1)) \right]$.
The optimization objective is to find the optimal parameters $\theta^*$ that minimize the loss function across the noisy condition, $\theta^* = \arg\min_{\theta} \mathbb{E}_{(\mathbf{x}_i, y_i) } \left[\mathcal{L}(y_i, \mathcal{M}_\theta(\mathcal{N}(\rho_i)))\right]$, where $\mathcal{M}_\theta(\mathcal{N}(\rho_i))$ represents the QNN's output under the influence of quantum noise.

 While QNNs rely on quantum circuits to process data, the next model, the VQC, combines quantum circuits with classical optimization to solve classification problems.
 
 \vspace{-4 mm}
\subsection{VQC}

The VQC is a hybrid algorithm that utilizes parameterized quantum circuits (PQC) to transform input data and employs classical algorithms to optimize the parameters \cite{jager2023universal, havlivcek2019supervised}.
VQC encodes data into quantum states and processes them using a variational circuit \(U(\theta)\). Measurements on the output states provide information used for classification. The parameters \(\theta\) are adjusted to minimize a loss function, improving the classifier's performance.
Quantum noise impacts the VQC by altering the quantum states during processing and affecting measurement results. This leads to errors in the computed expectation values used for updating the parameters, potentially hindering the convergence and accuracy of the classifier.

The first step in the VQC is to encode the classical input data $\mathbf{x}_i$ into a quantum state $\rho_i$. 
The PQC is defined by a set of parameters $\theta$, which are adjusted during the training process. The action of the PQC on the quantum state $\rho_i$ is described by,
$\mathbf{U}(\theta): \rho_i \mapsto \hat{\rho_{i}} = \mathbf{U}(\theta) \rho_i \mathbf{U}^\dagger(\theta)$. Where $\mathbf{U}(\theta)$ is a unitary operator generated by the PQC with parameters $\theta$, and $\hat{\rho_i}$ is the quantum state after applying the PQC. A measurement is then performed, using a set of Pauli operators $\{\sigma_j\}$, and the expectation value of these operators provides the output.
$\hat{y}_i = \text{sign}\left(\sum_{j} w_j \langle \hat{\rho_i} | \sigma_j | \hat{\rho_i} \rangle \right)$, where $w_j$ are weights associated with the measurement outcomes and $\text{sign}(\cdot)$ function determines the predicted class label $\hat{y}_i$.
The PQC then processes this noisy state $\rho_i'$, and the measurement is performed on the transformed noisy state, $\rho_i' = U(\theta) \rho_i' U^\dagger(\theta)$.
The training process of the VQC involves minimizing a loss function $\mathcal{L}(\theta)$ that quantifies the difference between the predicted labels $\hat{y}_i$ and the true labels $y_i$. In the presence of quantum noise, this optimization problem can be expressed as:
$\theta^* = \arg\min_{\theta} \mathbb{E}_{(\mathbf{x}_i, y_i)} \left[\mathcal{L}\left(y_i, \text{sign}\left(\sum_{j} w_j \langle \rho_i' | \sigma_j | \rho_i' \rangle \right)\right)\right]$.

In this section, we have modeled the impact of quantum noise on four fundamental QML algorithms. We have highlighted how quantum noise affects each algorithm's performance by providing intuitive explanations and mathematical formulations.

\vspace{-4 mm}
\section{Result and Analysis}

In this study, we utilized the \textit{democoding vs intergenomic} dataset from the Genomic Benchmarks project \cite{grevsova2023genomic}, designed to classify genomic sequences as either protein-coding or non-coding. The dataset contains 100,000 sequences, equally split between the two classes, each with a uniform length of 200 base pairs. To gain practical insights into the application of NISQ devices for genome sequence classification, we employ IBM's \textit{AerSimulator} to replicate the realistic noisy conditions inherent in IBM's quantum computers.
Additionally, we implemented four distinct QML algorithms, as described in the previous section, using IBM's Qiskit toolkit \cite{javadi2024quantum}. These implementations followed standard configurations, with all parameters set to their default values. This approach provides a consistent baseline for evaluating the performance of these models under varying conditions, which will be analyzed in subsequent sections for genome sequence classification. 

We used a subset of the dataset for training and testing the QML models, converting genome sequences into numerical format through text vectorization. To reduce computational complexity, we applied Principal Component Analysis (PCA) \cite{singh2024independent}, reducing the data to four dimensions, corresponding to the four qubits used in this study. We then encoded the reduced data into quantum states using the feature maps previously discussed.
\vspace{-3 mm}
\subsection{Quantum State Preparation}
Quantum noise significantly impacts QML, particularly during quantum state preparation, where it can disrupt state fidelity and model performance.
The impact varies across different feature maps, showing distinct levels of sensitivity to noise and state degradation.

\begin{figure}[h!]
    \centering
    \includegraphics[scale=0.5]{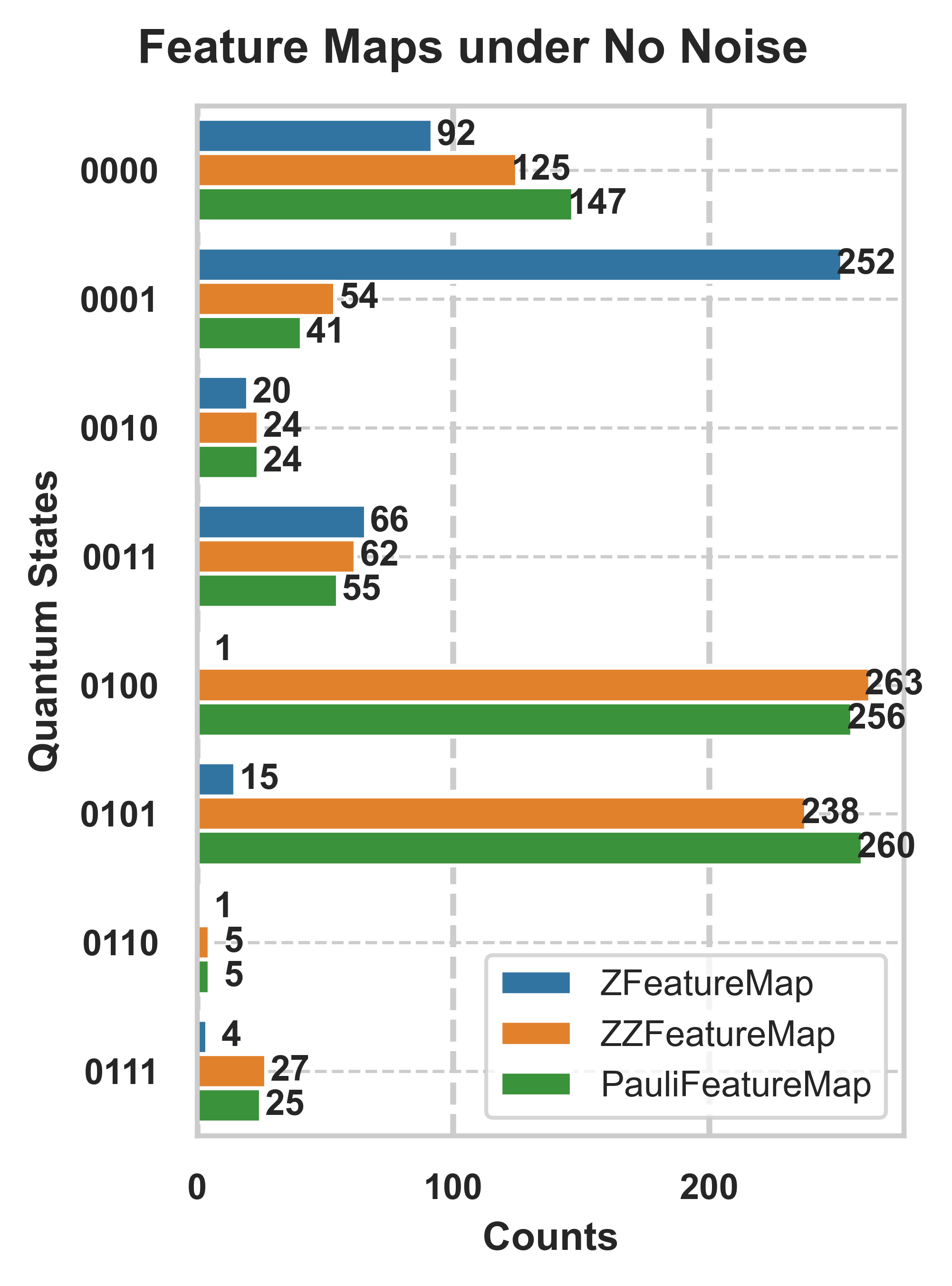}
    \caption{Output state under no noise conditions.}
    \label{fig:no_noise}
    \vspace{-4 mm}
\end{figure}

In the absence of noise as shown in Fig. \ref{fig:no_noise}, ZFeatureMap, ZZFeatureMap, and PauliFeatureMap present distinct patterns in the output state distributions, primarily concentrating counts around a few high-probability states. For instance, in the ZFeatureMap, states such as $|0000\rangle$ and $|0001\rangle$, indicate that these quantum states capture most of the information encoded by the feature map. The ZZFeatureMap shows a more evenly spread distribution across high-probability states, with states like $|0100\rangle$, and $|0101\rangle$ being highly frequent. The PauliFeatureMap exhibits a mix of both concentrated and dispersed states, with $|0101\rangle$ and $|0100\rangle$ having the highest counts.
These results establish a baseline for comparison against noisy environments. The high counts in certain states reflect that these states are strongly aligned with the encoded genome features, and the overall feature map performance is stable, with minimal fluctuations across the quantum states.

\begin{figure}[h!]
    \centering
    \includegraphics[width=0.7\columnwidth]{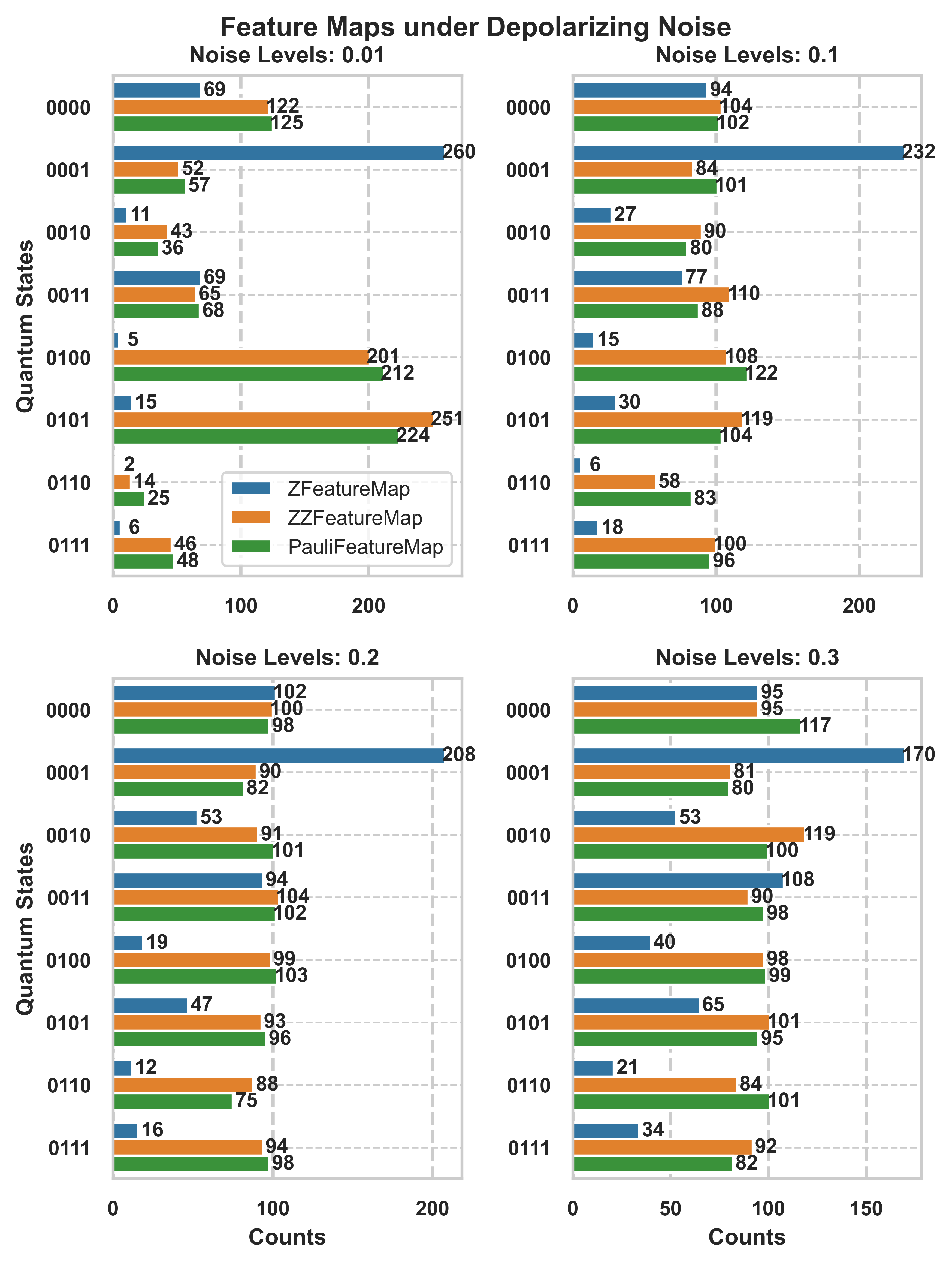}
    \caption{Output state under depolarizing noise.}
    \label{fig:dep_featuremap}
    \vspace{-4 mm}
\end{figure}

Depolarizing noise introduces random errors into the quantum circuit, affecting both single and two-qubit gates. As shown in Fig. \ref{fig:dep_featuremap}, the error rate increases from $0.01$ to $0.3$, the output state distributions show notable shifts. In the ZFeatureMap, the counts for the dominant states decrease, while the counts for lower-probability states increase. This pattern indicates that depolarizing noise effectively randomizes the quantum states, leading to a more uniform distribution as noise intensifies.
For the ZZFeatureMap, depolarizing noise reduces the concentration of counts around states like $|0100\rangle$ and $|0101\rangle$, with the state distribution becoming more uniform as noise increases. Similarly, the PauliFeatureMap demonstrates a decline in the dominance of high-probability states (e.g., $|0101\rangle$, $|0100\rangle$), with noise dispersing the counts across more states. This dispersion implies that the quantum circuit's ability to preserve the encoded information deteriorates with increasing depolarizing noise.
The impact of depolarizing noise is significant because it leads to a reduction in the model's ability to retain encoded features, thus reducing the quantum circuit’s capacity to classify genome sequences effectively.
\begin{figure}[h!]
    \centering
    \includegraphics[width=0.7\columnwidth]{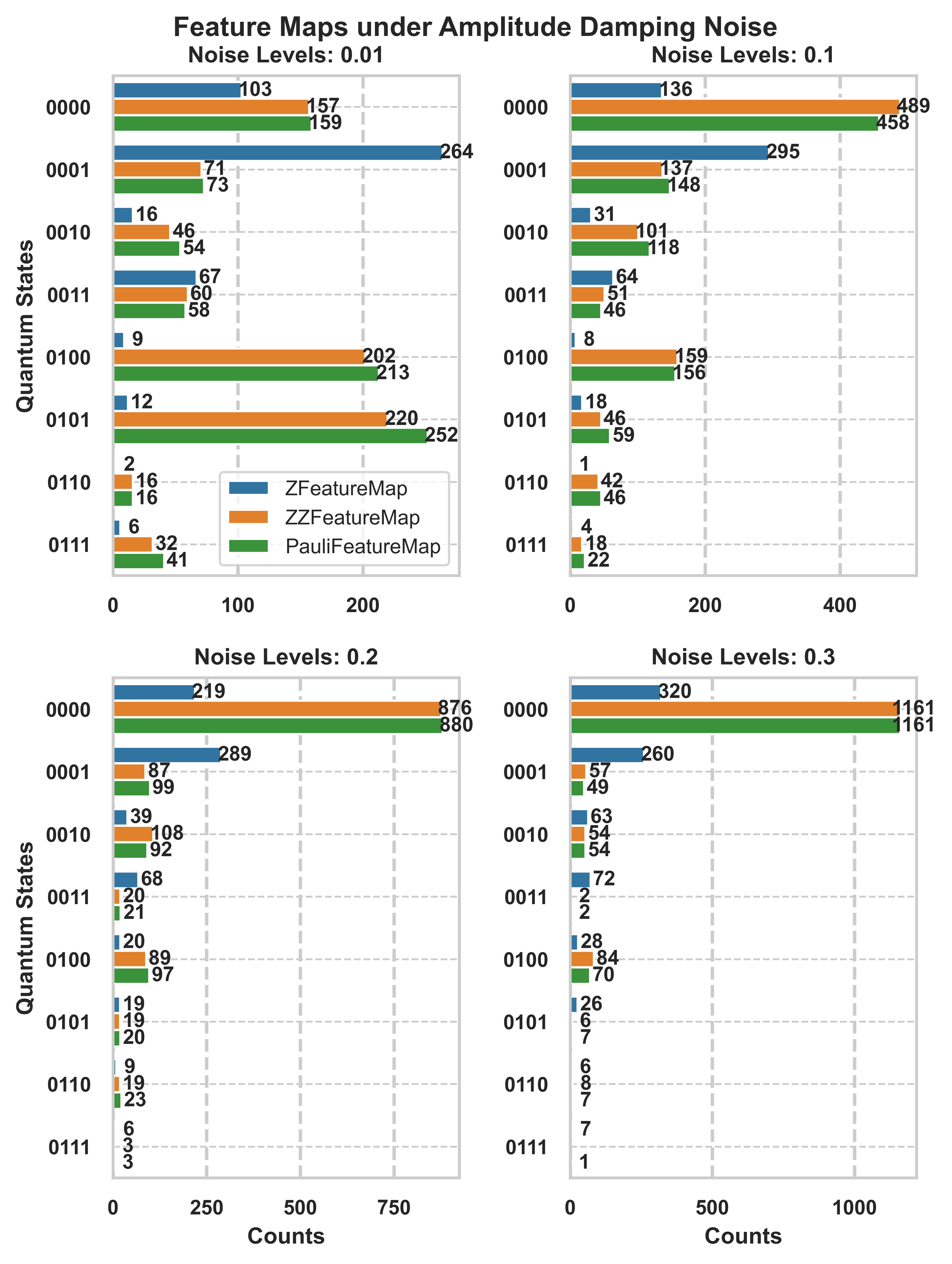}
    \caption{Output state under amplitude damping noise.}\vspace{-4 mm}
    \label{fig:amp_featuremap}
    
\end{figure}

As we discussed earlier, amplitude damping noise models energy loss, where quantum states decay to lower energy levels. In Fig. \ref{fig:amp_featuremap} at an error rate of $0.01$, the ZFeatureMap shows a modest reduction in the counts for high-probability states, but as the error rate increases to $0.3$, these counts decline sharply. States like $|0000\rangle$ and $|0001\rangle$ begin to dominate, which reflects the impact of amplitude damping pushing the system toward lower-energy states.
The ZZFeatureMap and PauliFeatureMap exhibit similar trends, with high-probability states losing their dominance as noise increases. In the ZZFeatureMap, states suffer significant reductions in counts at higher noise levels, while in the PauliFeatureMap, dominant states see a marked decrease in favor of lower-energy states like $|0000\rangle$.
Amplitude damping noise has a particularly strong effect on quantum circuits because it preferentially collapses the system to lower-energy states, severely impacting the quantum feature maps' ability to represent complex data.

\begin{figure}[h!]
    \centering
    \includegraphics[width=0.7\columnwidth]{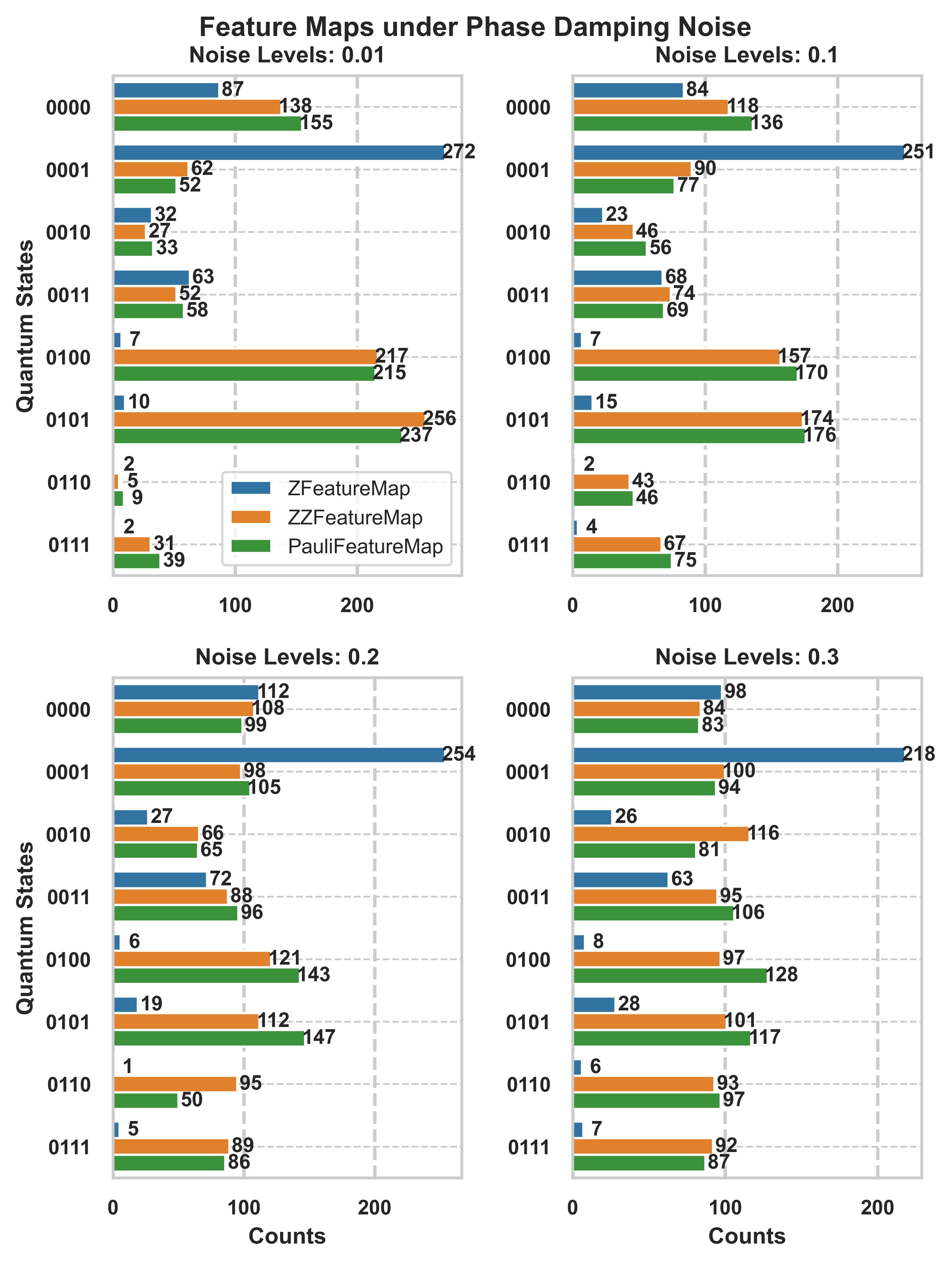}
    \caption{Output state under phase damping noise.}
    \label{fig:pahsedam_featuremap}
    \vspace{-4 mm}
\end{figure}

Phase damping noise affects the coherence of quantum states without energy loss. As shown in Fig. \ref{fig:pahsedam_featuremap} in the ZFeatureMap, the impact of phase damping is visible at a noise level of $0.01$, where the counts for key states slightly decrease. As the noise level increases to $0.3$, the dispersion of counts across states becomes more pronounced. High-probability states see reduced counts, while other states, like $|0000\rangle$ and $|0011\rangle$, gain prominence.
The ZZFeatureMap shows a gradual decline in coherence with increasing phase damping noise. High-probability states see decreasing counts, while previously low-probability states start to gain counts. The PauliFeatureMap follows a similar trend, with dominant states losing their strength as noise increases.
Phase damping noise impacts the circuit's ability to maintain quantum coherence, leading to a loss of information in the encoded feature maps. This loss makes it harder for the QML models to classify genome sequences effectively, as the coherence necessary to preserve the quantum state's information is degraded.
\begin{figure}[h!]
    \centering
    \includegraphics[width=0.7\columnwidth]{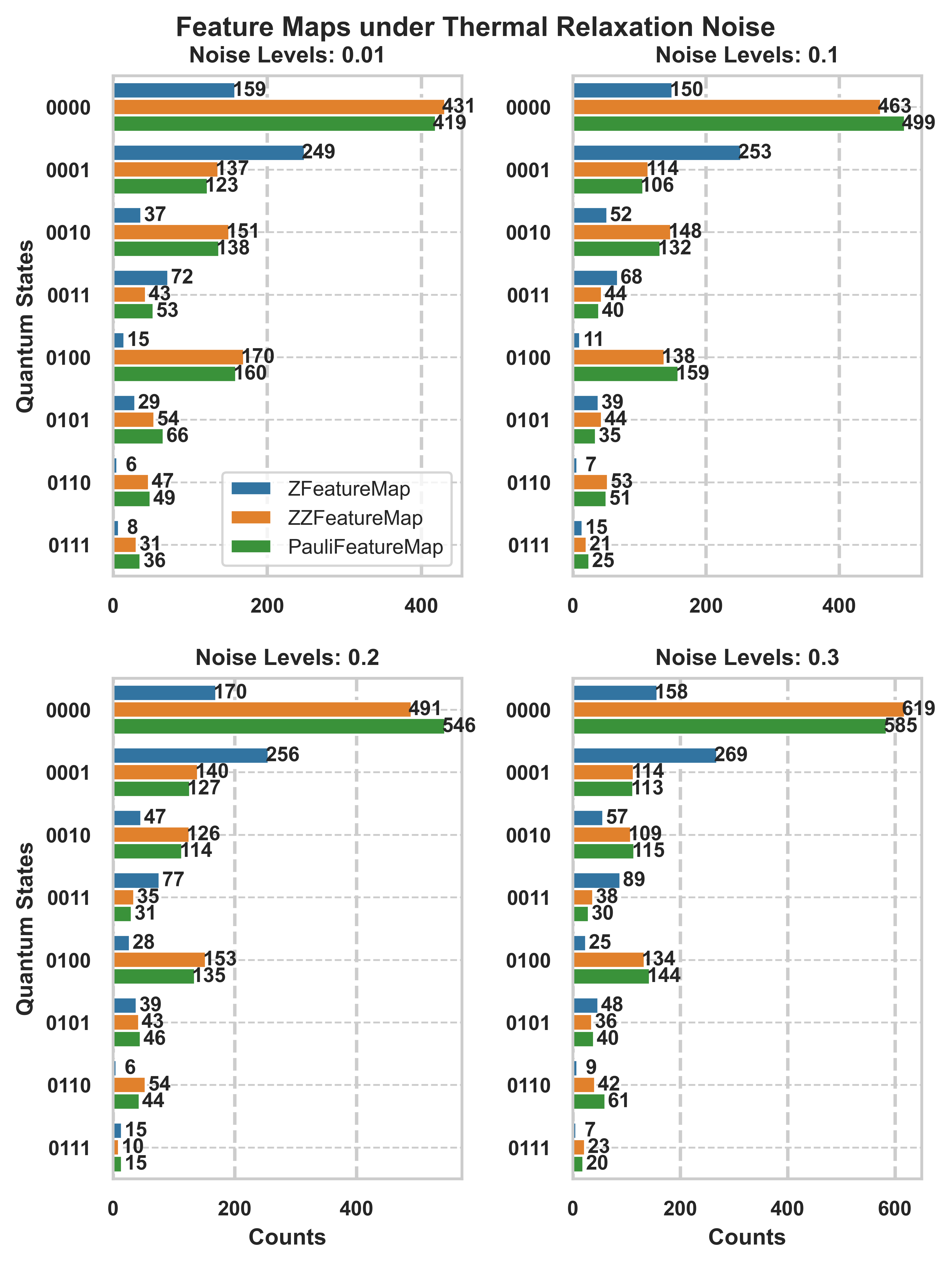}
    \caption{Output state under thermal relaxation noise.}
    \label{fig:thermal_featuremap}
    \vspace{-4 mm}
\end{figure}

Thermal relaxation noise models both amplitude damping and phase damping together. As shown in Fig.  \ref{fig:thermal_featuremap}, at low noise levels $(0.01)$, the ZFeatureMap shows slight decreases in the counts for high-probability states. However, as the noise level increases, these states' counts decrease significantly, with low-energy states becoming dominant.
The ZZFeatureMap experiences a similar trend, with high-probability states losing their dominance as noise increases. In the PauliFeatureMap, the impact of thermal relaxation is pronounced at higher noise levels, with high-probability states seeing a substantial drop in counts.
Thermal relaxation noise severely impacts the quantum circuits by combining the effects of energy loss and coherence loss. This combination makes it challenging for the QML models to maintain accurate genome sequence classifications, as the circuits tend to collapse into lower-energy states, reducing the efficacy of the feature maps.
\begin{figure}[h!]
    \centering
    \includegraphics[width=0.7\columnwidth]{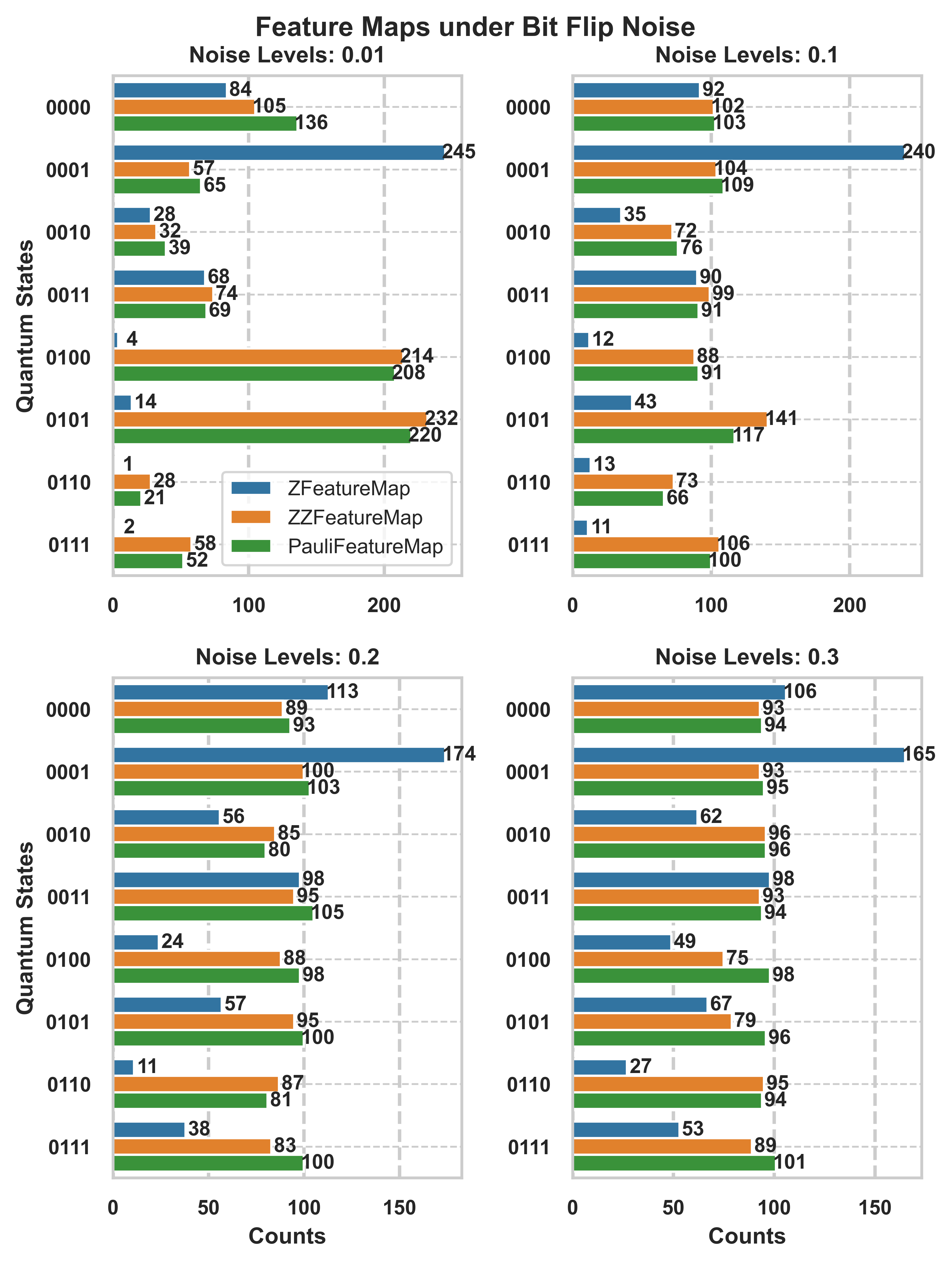}
    \caption{Output state under bit flip noise.}
    \label{fig:bit_featuremap}
    \vspace{-4 mm}
\end{figure}

\begin{figure}[h!]
    \centering
    \includegraphics[width=0.7\columnwidth]{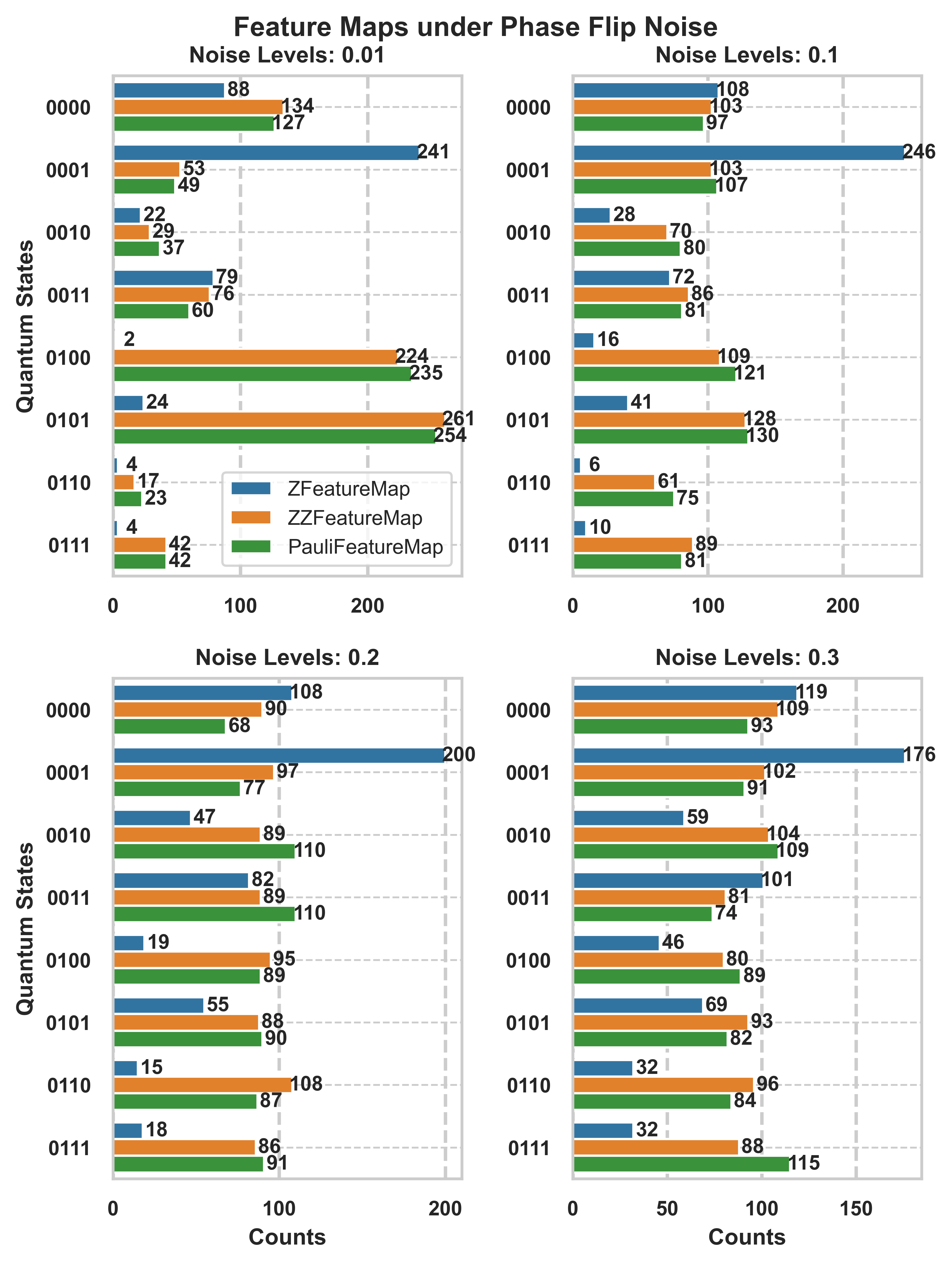}
    \caption{Output state under phase flip noise.}
    \label{fig:phasef_featuremap}
    \vspace{-4 mm}
\end{figure}

Bit flip and phase flip noise models discrete errors where quantum states are flipped between $|0\rangle$ and $|1\rangle$ or undergo a phase change. As shown in Figs. \ref{fig:bit_featuremap} and \ref{fig:phasef_featuremap} for both noise types, the ZFeatureMap shows a gradual reduction in the counts for high-probability states as noise levels increase. In the case of bit flip noise, states are flipped, leading to more uniform distributions across the quantum states.
The ZZFeatureMap and PauliFeatureMap exhibit similar trends, with high-probability states losing their dominance as noise levels rise. However, bit flip noise has a more pronounced impact on the PauliFeatureMap, where dominant states experience significant flipping, resulting in increased counts for previously low-probability states.
Bit flip and phase flip noise primarily disrupts the encoded information by flipping quantum states. This disruption results in a loss of the structured patterns that the feature maps aim to preserve, degrading the performance of the QML models in noisy environments.

The experiments reveal that each type of quantum feature map responds differently to noise, highlighting the varying resilience of these encoding strategies. Without noise, the feature maps try to effectively encode the genome sequences, concentrating counts in a few high-probability quantum states that capture the primary features of the data. However, as noise levels increase, the counts become more dispersed, indicating a loss of information.

This degradation is significant because it shows how noise impacts the ability of these QML models to encode, process, and classify complex biological data. By understanding how different feature maps behave under various noise types, one can select or design feature maps that are more resilient to specific noise environments, potentially improving the accuracy and stability of QML models. Another significant observation from these results is that these feature mapping techniques lack the ability to preserve the sequential information of the genomics data. These techniques need more qubits to encode the data, which significantly reduces the scalability aspect of these techniques when data have an extensive feature set. 
 \vspace{-4mm }
\subsection{Training and Test Performance}
We now present the training and testing accuracies of various discussed QML models under different quantum noise levels using encoding techniques. We analyze the results for error rates of $0.01$, $0.1$, $0.2$, and $0.3$, along with varying shots. The general trend observed indicates that while training accuracy remains relatively stable across varying noise levels, testing accuracy experiences a significant decrease.

\subsubsection{Amplitude Damping Noise}

\begin{figure}[ht!]
    \centering
    \begin{minipage}[b]{0.49\textwidth}
        \centering
        \includegraphics[width=\textwidth]{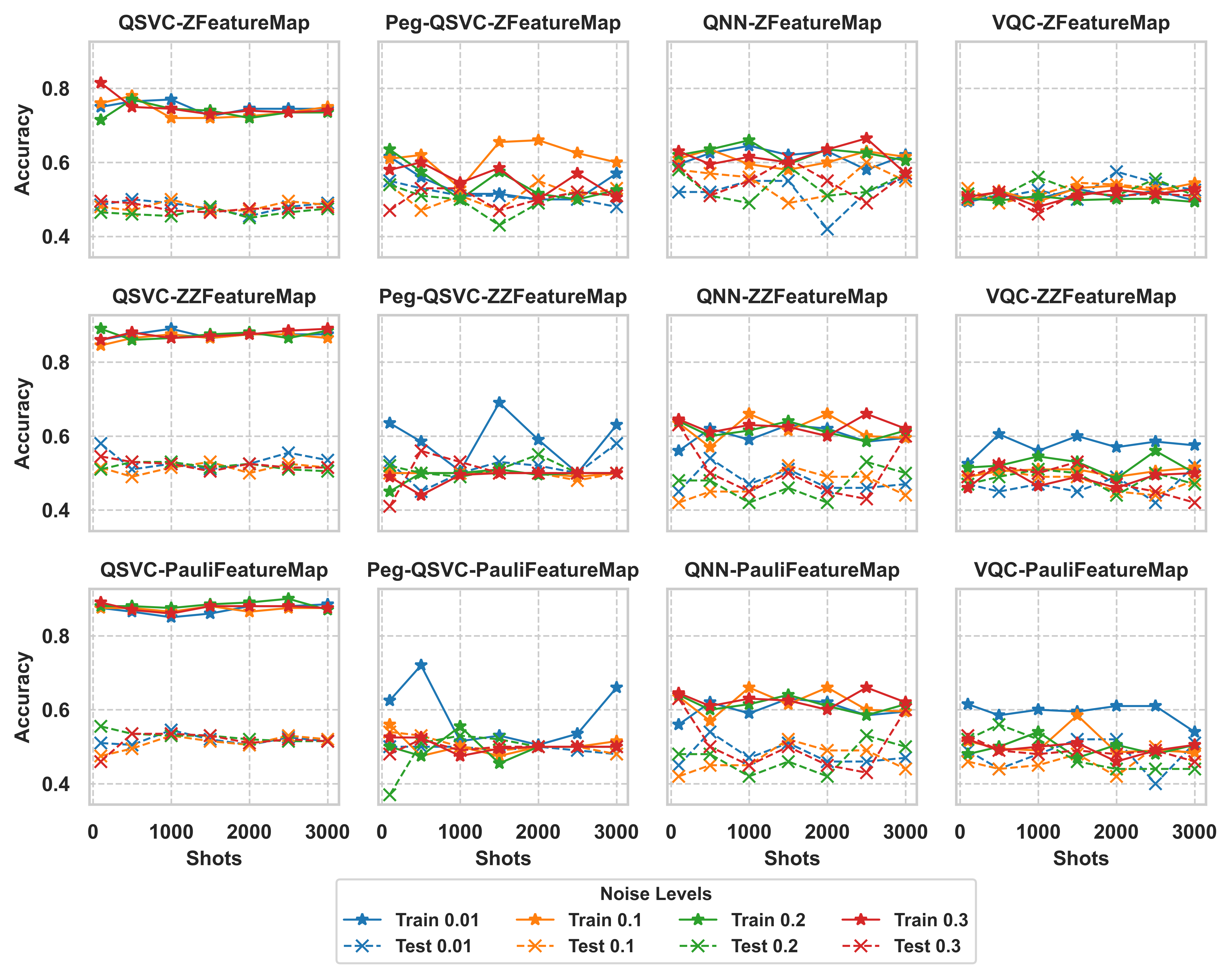}
        \caption{Train and test accuracy under amplitude damping noise.}
        \label{fig:amp_damp_acc}
    \end{minipage}
    \hfill
    \begin{minipage}[b]{0.49\textwidth}
        \centering
        \includegraphics[width=\textwidth]{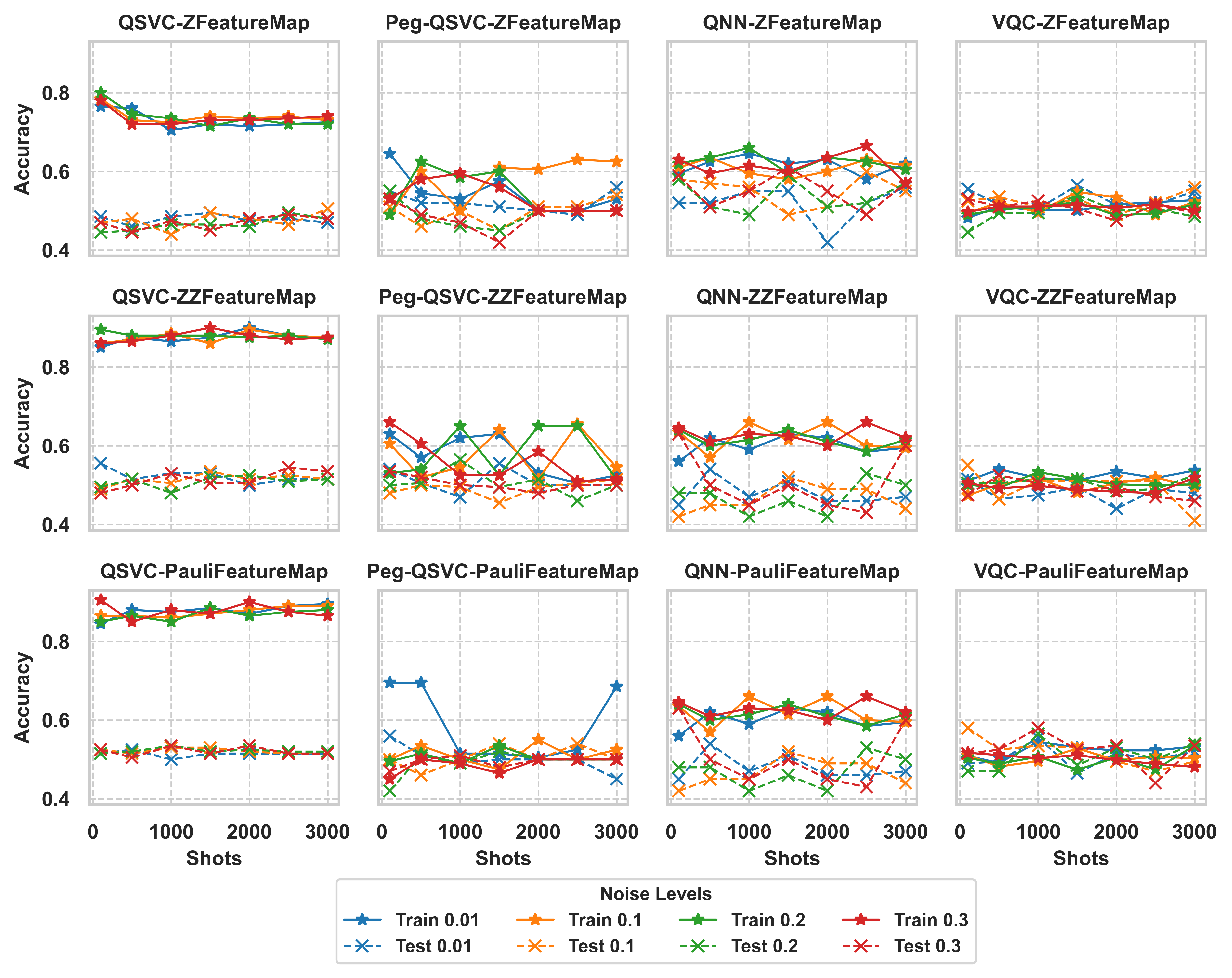}
        \caption{Train and test accuracy under depolarizing noise.} \vspace{-7mm }
        \label{fig:dep_acc}
    \end{minipage}
    \vspace{-2 mm}
\end{figure}

Fig. ~\ref{fig:amp_damp_acc}  illustrates the impact of amplitude damping noise on the performance of various QML models using different feature maps. Across all models and feature maps, training accuracy remains relatively stable regardless of noise levels and the number of shots, indicating robustness in fitting training data despite amplitude damping noise. However, testing accuracy varies more significantly, particularly at higher noise levels (e.g., $0.2$ and $0.3$), as increased decoherence degrades the models' generalization capabilities.

ZFeatureMap shows the most stability, especially in QSVC and Peg-QSVC, with minimal testing accuracy degradation, due to its encoding along the Z-axis, making it less sensitive to amplitude damping. In contrast, ZZFeatureMap introduces entanglement, leading to more testing accuracy fluctuations, particularly in Peg-QSVC, due to the vulnerability of entangled states to noise. PauliFeatureMap exhibits stable training accuracy, but testing accuracy is more sensitive to noise, particularly in Peg-QSVC and QNN.

QSVC performs best across noise levels, especially with ZFeatureMap and ZZFeatureMap, due to its resilient kernel methods. Peg-QSVC and QNN are more sensitive to noise, with Peg-QSVC showing variability across all feature maps and QNN also being particularly affected by noise in ZZFeatureMap and PauliFeatureMap.

\subsubsection{Depolarizing Noise}
Fig. ~\ref{fig:dep_acc} shows the accuracy of various QML models using different feature maps under depolarizing noise. In this training, accuracy remains stable and testing accuracy varies significantly, especially at higher noise levels, as the uniform probability of error introduced by depolarizing noise degrades generalization more than amplitude damping noise.

Performance under the ZFeatureMap is generally stable, with slight testing accuracy degradation in QSVC and Peg-QSVC as noise increases. QNN and VQC show more fluctuations, indicating their higher sensitivity to random errors. The ZZFeatureMap, due to entanglement, makes models more vulnerable to noise, particularly Peg-QSVC and QNN, which experience significant testing accuracy drops. PauliFeatureMap maintains stable training accuracy but shows pronounced declines in testing accuracy, especially in Peg-QSVC and QNN, due to increased sensitivity to random errors.

QSVC shows resilience across all feature maps, with minimal testing accuracy loss, likely due to its robust kernel methods. Peg-QSVC displays significant variability, particularly with the ZZFeatureMap and PauliFeatureMap, possibly due to the Pegasos algorithm's iterative nature. QNN exhibits notable sensitivity, especially with ZZFeatureMap, due to the complexity of neural networks and entanglement. VQC demonstrates moderate noise sensitivity, with noise-induced optimization challenges, particularly with the PauliFeatureMap.

\subsubsection{Phase Damping/Dephasing Noise}
\begin{figure}[ht!]
    \centering
    \begin{minipage}[b]{0.49\textwidth}
        \centering
        \includegraphics[width=\textwidth]{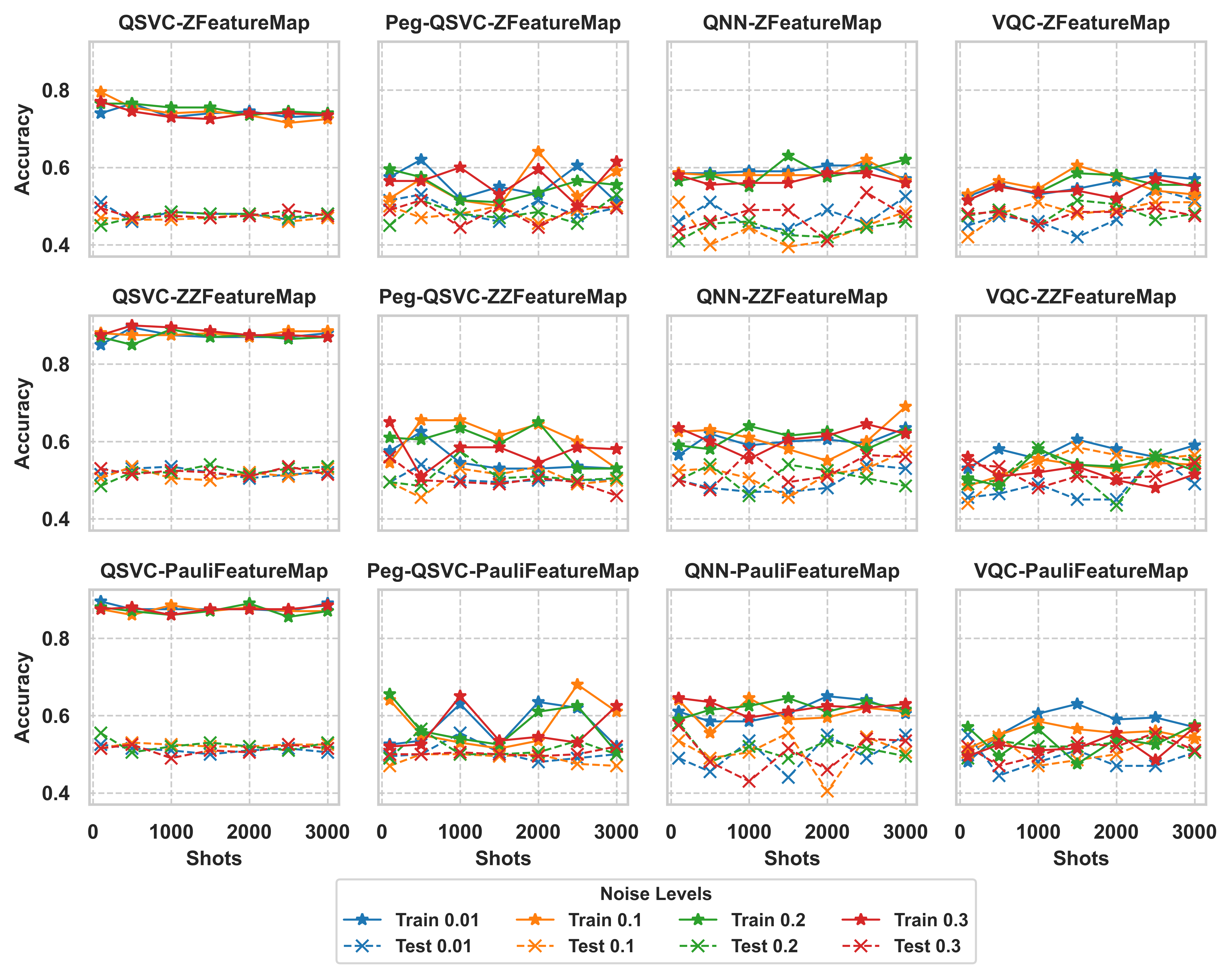}
        \caption{Train and test accuracy under dephase damping noise.}
        \label{fig:Phase_damp_acc}
    \end{minipage}
    \hfill
    \begin{minipage}[b]{0.49\textwidth}
        \centering
        \includegraphics[width=\textwidth]{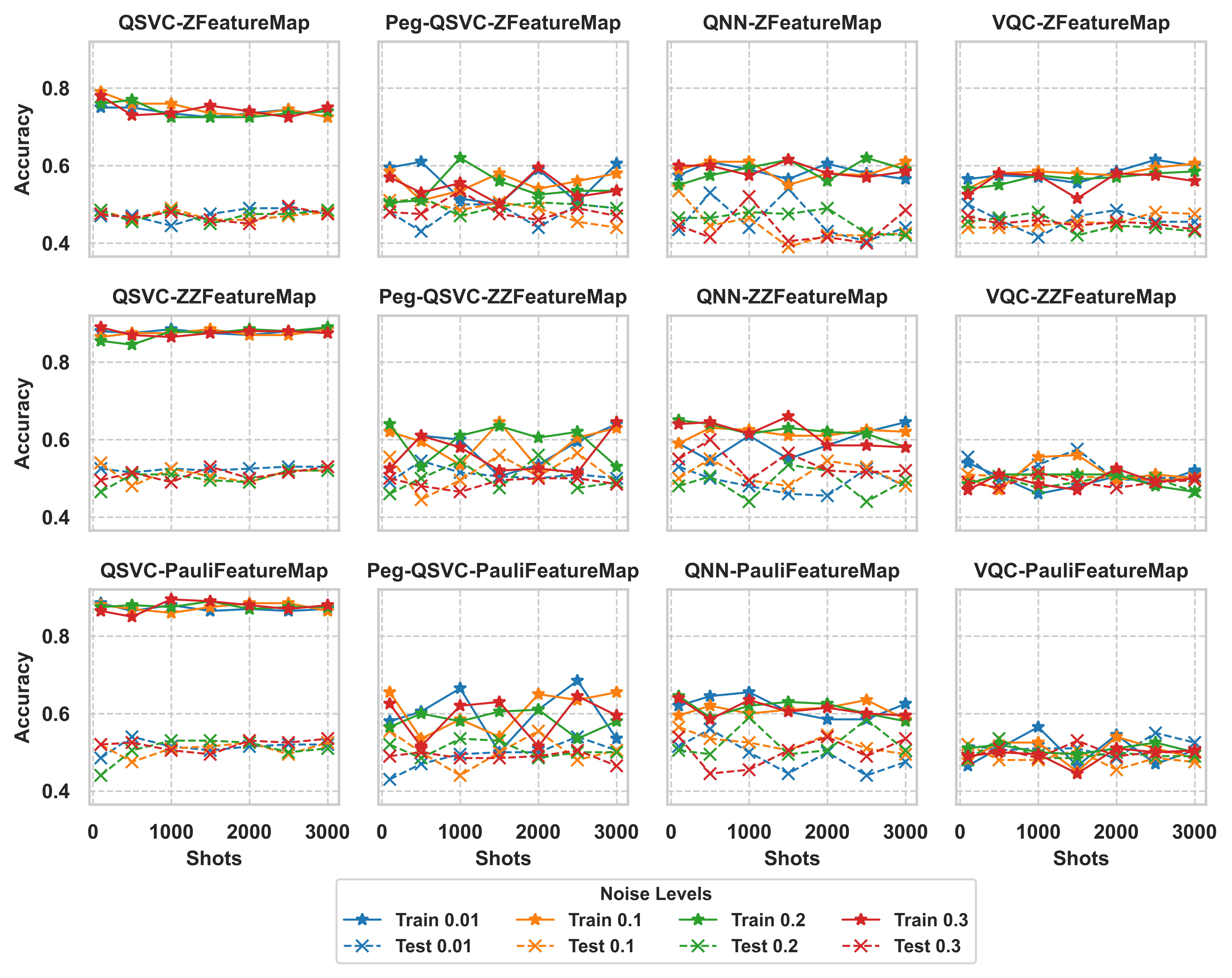}
        \caption{Train and test accuracy under thermal relaxation noise.} \vspace{-7mm }
        \label{fig:Thermal_Relaxation_noise}
    \end{minipage}
    \vspace{-4 mm}
\end{figure}

Fig.~\ref{fig:Phase_damp_acc} presents the accuracy of various QML models using different feature maps under phase damping noise, which represents the loss of coherence without energy dissipation and affects phase relationships between quantum states crucial for quantum algorithms.

ZFeatureMap demonstrates relative robustness, with QSVC maintaining stable testing accuracy even at higher noise levels. In contrast, Peg-QSVC and QNN show more fluctuations, suggesting that model complexity influences resilience to phase damping. ZZFeatureMap, which introduces entanglement, is highly sensitive to phase damping noise, especially in the Peg-QSVC, where testing accuracy varies significantly. PauliFeatureMap also exhibits stable training accuracy but declines in testing accuracy, particularly in Peg-QSVC and QNN, due to its reliance on phase relationships.

QSVC shows the highest resilience across feature maps, likely due to its lesser dependence on phase coherence. Peg-QSVC is notably sensitive to phase damping, especially with ZZFeatureMap and PauliFeatureMap, as the iterative approximation process amplifies phase decoherence effects. QNN also shows considerable sensitivity, particularly with ZZFeatureMap, where disrupted phase relationships significantly impact performance. VQC shows moderate sensitivity, with some stability in the ZFeatureMap, but noise still affects testing accuracy due to challenges in variational optimization under phase decoherence.

\subsubsection{Thermal Relaxation Noise}
Fig. ~\ref{fig:Thermal_Relaxation_noise} shows the accuracy of various QML models using different feature maps under thermal relaxation noise. Testing accuracy shows significant sensitivity to thermal relaxation noise, particularly at higher noise levels, suggesting that coherence loss impacts generalization and performance on unseen data.
ZFeatureMap shows relatively stable performance, with the QSVC maintaining high testing accuracy even at higher noise levels, while Peg-QSVC and QNN exhibit more variability, indicating model complexity affects robustness. ZZFeatureMap, which involves entanglement, is more sensitive to noise, especially in Peg-QSVC and QNN, where significant testing accuracy fluctuations occur due to susceptibility to energy dissipation and dephasing. PauliFeatureMap also shows stable training accuracy but declines in testing accuracy as noise increases, particularly in Peg-QSVC and QNN, due to its reliance on phase and amplitude coherence.

QSVC demonstrates resilience across feature maps, especially with ZFeatureMap and ZZFeatureMap, likely due to kernel methods that focus on decision boundaries less affected by noise. Peg-QSVC shows significant variability, particularly with ZZFeatureMap and PauliFeatureMap, as the iterative Pegasos algorithm amplifies noise effects. QNN exhibits noticeable fluctuations, especially with ZZFeatureMap, due to its complex structure and entanglement vulnerability. VQC displays moderate sensitivity across all feature maps, with declines in testing accuracy due to noise-affected parameter optimization, particularly in complex feature maps.

\begin{figure}[ht!]
    \centering
    \begin{minipage}[b]{0.49\textwidth}
        \centering
        \includegraphics[width=\textwidth]{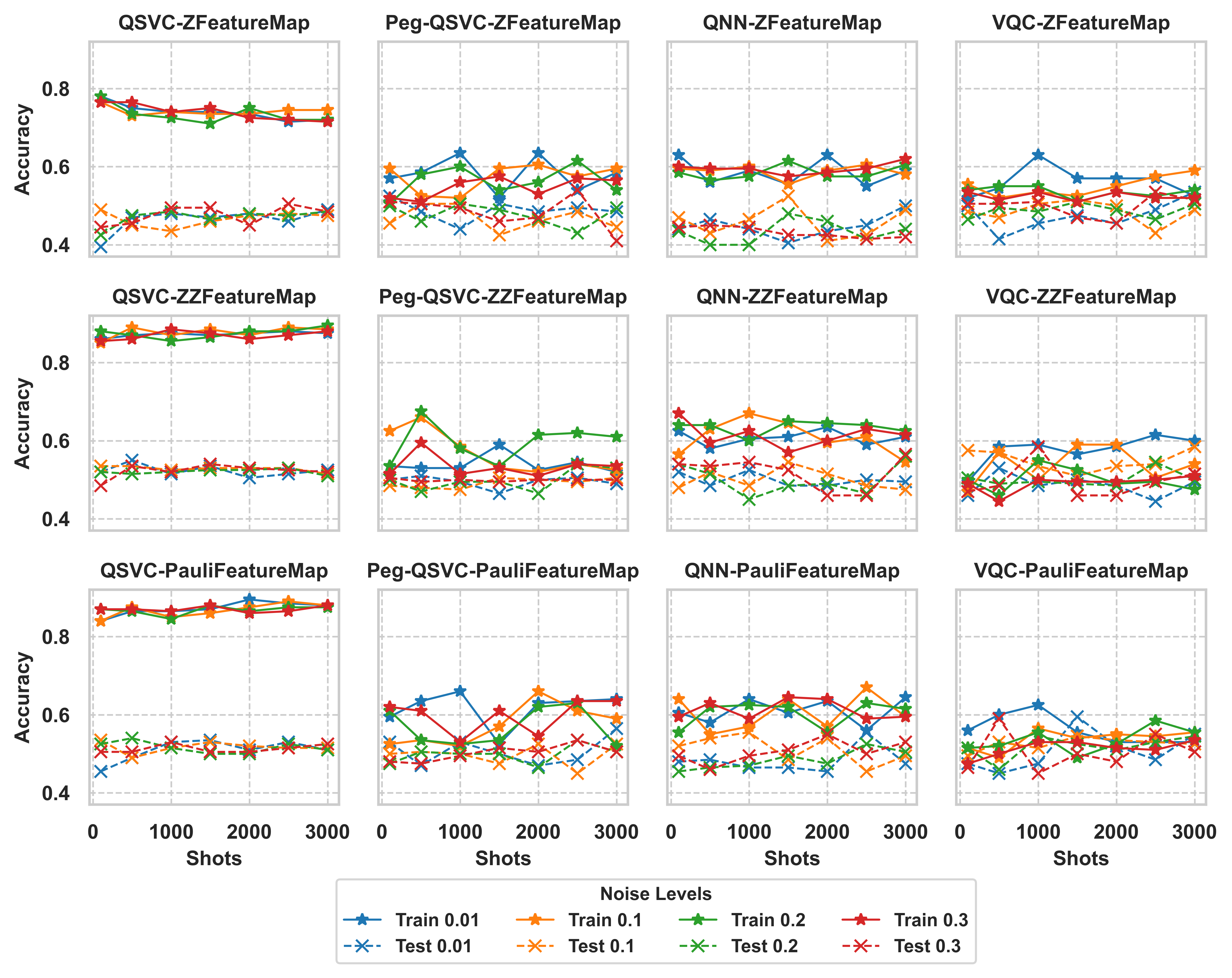}
        \caption{Train and test accuracy under bit flip noise.}
        \label{fig:Bit_Flip_noise}
    \end{minipage}
    \hfill
    \begin{minipage}[b]{0.49\textwidth}
        \centering
        \includegraphics[width=\textwidth]{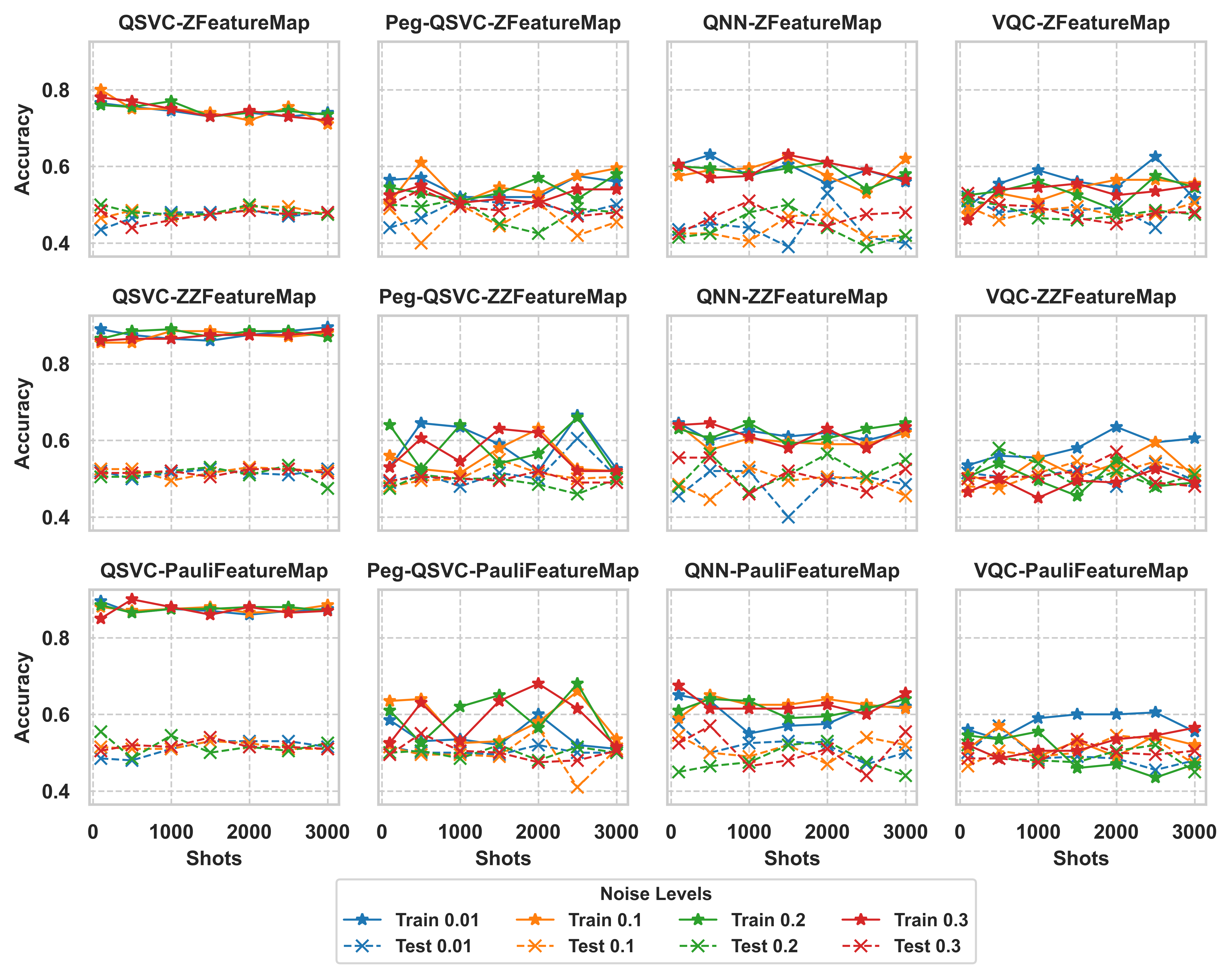}
        \caption{Train and test accuracy under phase flip noise.}
        \vspace{-5mm }
        \label{fig:Phase_Flip_noise}
    \end{minipage}
    
\end{figure}

\subsubsection{Bit Flip Noise}
 Fig.~\ref{fig:Bit_Flip_noise}  presents the accuracy of various QML models using different feature maps under bit flip noise. ZFeatureMap exhibits resilience, with QSVC maintaining stable testing accuracy even at higher noise levels. However, Peg-QSVC and QNN show more fluctuations, indicating that while ZFeatureMap mitigates some bit flip effects, model complexity influences robustness. ZZFeatureMap is more sensitive due to qubit entanglement, with Peg-QSVC and QNN experiencing significant testing accuracy drops. PauliFeatureMap, though stable in training accuracy, shows pronounced testing accuracy declines, especially in Peg-QSVC and QNN, due to its reliance on specific state preparations that are vulnerable to bit flips.

QSVC demonstrates strong resilience, especially with ZFeatureMap and ZZFeatureMap, likely due to the kernel method's focus on decision boundaries, which are less sensitive to state flips. Peg-QSVC shows significant variability across all feature maps, particularly with ZZFeatureMap and PauliFeatureMap, as its iterative process is sensitive to cumulative bit flips. QNN shows considerable fluctuations, especially with ZZFeatureMap, due to its complex structure and entanglement. VQC displays moderate sensitivity, with bit flip noise impacting its optimization process, particularly with complex feature maps.

\subsubsection{Phase Flip Noise}
Fig.~\ref{fig:Phase_Flip_noise} presents the accuracy of various QML models using different feature maps under phase flip noise. Phase flip noise inverts the phase of a qubit while leaving the amplitude unchanged, significantly impacting quantum algorithms that rely on phase coherence. ZFeatureMap demonstrates reasonable robustness, with the QSVC maintaining stable testing accuracy even at higher noise levels. In contrast, Peg-QSVC and QNN show more fluctuations, indicating that while ZFeatureMap mitigates some phase flip effects, model complexity still influences resilience. ZZFeatureMap, involving entanglement, shows heightened sensitivity to phase flip noise, with Peg-QSVC and QNN exhibiting significant testing accuracy drops. PauliFeatureMap also shows stable training accuracy but pronounced declines in testing accuracy, especially in Peg-QSVC and QNN, due to its reliance on phase coherence.

QSVC shows strong resilience, particularly with the ZFeatureMap and ZZFeatureMap, likely due to the kernel method's focus on decision boundaries less sensitive to phase disturbances. Peg-QSVC shows significant variability across all feature maps, especially with ZZFeatureMap and PauliFeatureMap, as its iterative process is sensitive to cumulative phase errors. QNN shows considerable fluctuations, especially with ZZFeatureMap, due to its complex structure and entanglement. VQC displays moderate sensitivity across feature maps, with optimization challenges due to phase flips, particularly in ZZFeatureMap and PauliFeatureMap.

\subsubsection{Impact of Noise Level}
Across different noise models, increasing noise levels significantly affects the performance of QML algorithms, especially in testing accuracy. At low noise levels $(0.01)$, models generally maintain high training and testing precision, showing resilience in all the noise types discussed. As noise levels increase to moderate $(0.1-0.2)$, a noticeable decrease in testing accuracy is observed, particularly in the Peg-QSVC and QNN models. The impact is more severe with complex quantum feature maps such as ZZFeatureMap and PauliFeatureMap, highlighting their sensitivity to noise. At high noise levels $(0.3)$, all models experience substantial accuracy drops, with QSVC using ZFeatureMap showing the highest resilience. In contrast, models like Peg-QSVC and QNN struggle significantly, especially when exposed to phase decoherence and state-flipping effects, suggesting the need for noise mitigation strategies.

\section{Conclusion}\label{sec-conclusion}
This study examined how quantum noise affects the performance of QML algorithms in genome sequence classification. While QML models like QSVC showed potential, their performance was significantly impacted by quantum noise in NISQ devices. QSVC demonstrated greater robustness across various noise levels and feature maps compared to the more sensitive Peg-QSVC and QNN. A key insight was that the sequential nature of genomic data requires specialized feature maps to preserve positional and sequential structures during quantum encoding. The ZFeatureMap was resilient to noise but failed to capture complex dependencies within sequence data. In contrast, the ZZFeatureMap and PauliFeatureMap incorporated entanglement and higher-order interactions, offering more expressive representations but were more susceptible to noise—particularly depolarizing and amplitude damping noise—which compromised the models' ability to differentiate complex sequence patterns.

\bibliographystyle{IEEEtran}
\bibliography{main}

\end{document}